\documentclass[preprint,review,11pt]{elsarticle}

\usepackage{graphicx}

\usepackage{amssymb}

\usepackage{caption}

\usepackage[top=3.0cm,left=2.4cm,right=2.4cm,bottom=3.0cm]{geometry}

\usepackage{enumitem} 
\setlist{itemsep=1ex plus0.2ex, leftmargin=*, align=left}

\makeatletter
\newcommand{\labitem}[2]{%
	\def\@itemlabel{\textbf{#1}}
	\item
	\def\@currentlabel{#1}\label{#2}}
\makeatother

\makeatletter
\newcommand{\headingitem}[1]{%
	\vspace{0.3cm}
	\def\@itemlabel{\textbf{#1}}
	\item
	\def\@currentlabel{#1}
	\addtocounter{enumi}{-1}
}
\makeatother

\usepackage{csvsimple}
\usepackage{multirow}
\usepackage{lipsum}

\usepackage{eurosym}
\usepackage{siunitx}

\DeclareSIUnit\eur{\officialeuro}
\DeclareSIUnit\M{M}
\DeclareSIUnit\k{k}

\def\sym#1{\ifmmode^{#1}\else\(^{#1}\)\fi}

\usepackage{etoolbox}
\robustify\bfseries

\usepackage{setspace} 
\usepackage{csquotes}
\usepackage{amsmath}
\usepackage{bm}

\usepackage{xspace}
\newcommand\ie{i.\,e.\xspace}
\newcommand\eg{e.\,g.\xspace}

\usepackage[amsmath,hyperref,framed]{ntheorem} 
\theoremstyle{plain}
\newtheorem{theorem}{Theorem}[section]

\theoremstyle{nonumberplain}
\theoremseparator{.}
\theoremheaderfont{\bfseries}
\theorembodyfont{\normalfont}
\theoremsymbol{$\blacksquare$}
\RequirePackage{amssymb}
\newtheorem{proof}{Proof}

\makeatletter
\let\copy@theorem@headerfont=\theorem@headerfont
\newcommand{\my@theorem@headerfont}{%
	\boldmath\copy@theorem@headerfont\unboldmath
}
\let\theorem@headerfont=\my@theorem@headerfont
\makeatother

\theoremstyle{nonumberplain}
\theoremseparator{.}
\newcommand{\expectation}{\mathbb{E}}

\newcommand{\argmax}{\operatornamewithlimits{arg \, max}}
\newcommand{\argmin}{\operatornamewithlimits{arg \, min}}

\usepackage[%
sort&compress
]{cleveref}
\usepackage{url}

\newcommand{\mathup}[1]{\mathrm{#1}}

\newcommand{\e}[1]{\mathup{e}^{#1}}

\newcommand{\abs}[1]{\left\lvert #1 \right\rvert}

\newcommand{\norm}[1]{\left\lVert#1\right\rVert}

\usepackage{algorithm}
\usepackage{algpseudocode}

\usepackage[usenames,dvipsnames]{xcolor}

\usepackage{colortbl}
\usepackage{tabularx}
\usepackage{booktabs}
\usepackage{collcell}

\usepackage{ragged2e}

\newcommand{\PreserveBackslash}[1]{\let\temp=\\#1\let\\=\temp}
\newcolumntype{v}[1]{>{\PreserveBackslash\RaggedRight\hspace{0pt}}p{#1}}

\usepackage{adjustbox}
\newcolumntype{Q}[2]{%
	>{\adjustbox{angle=#1,lap=\width-(#2)}\bgroup}%
	l%
	<{\egroup}%
}

\newcommand{\mcellt}[2][c]{%
	\begin{tabular}[t]{@{}#1@{}}#2\end{tabular}}
\newcommand{\lcellt}[2][l]{%
	\begin{tabular}[t]{@{}#1@{}}#2\end{tabular}}

\usepackage[
colorinlistoftodos,
textsize=footnotesize,
]{todonotes}

\makeatletter
\renewcommand{\fps@figure}{htb}         
\renewcommand{\fps@table}{htb}         
\makeatother 

\makeatletter
\newcommand\notsotiny{\@setfontsize\notsotiny{7.5}{8.5}}
\makeatother

\usepackage{float}
\usepackage{rotating}

\journal{Decision Support Systems}

\begin{document}
	
\begin{frontmatter}



\title{Forecasting remaining useful life: Interpretable deep learning approach via variational Bayesian inferences\\[1.2em]}


\author[ETH]{Mathias Kraus\corref{cor1}}
\ead{mathiaskraus@ethz.ch}

\author[ETH]{Stefan Feuerriegel}
\ead{sfeuerriegel@ethz.ch}

\address[ETH]{ETH Zurich, Weinbergstr. 56/58, 8092 Zurich, Switzerland}

\cortext[cor1]{Corresponding author.}

\begin{abstract}
Predicting the remaining useful life of machinery, infrastructure, or other equipment can facilitate preemptive maintenance decisions, whereby a failure is prevented through timely repair or replacement. This allows for a better decision support by considering the anticipated time-to-failure and thus promises to reduce costs. Here a common baseline may be derived by fitting a probability density function to past lifetimes and then utilizing the (conditional) expected remaining useful life as a prognostic. This approach finds widespread use in practice because of its high explanatory power. A more accurate alternative is promised by machine learning, where forecasts incorporate deterioration processes and environmental variables through sensor data. However, machine learning largely functions as a black-box method and its forecasts thus forfeit most of the desired interpretability. As our primary contribution, we propose a structured-effect neural network for predicting the remaining useful life which combines the favorable properties of both approaches: its key innovation is that it offers both a high accountability and the flexibility of deep learning. The parameters are estimated via variational Bayesian inferences. The different approaches are compared based on the actual time-to-failure for aircraft engines. This demonstrates the performance and superior interpretability of our method, while we finally discuss implications for decision support.
\end{abstract}

\begin{keyword}
Forecasting \sep Remaining useful life \sep Machine learning \sep Neural networks \sep Deep learning\\[0.3cm]
\noindent \textbf{Declarations of interest:} \emph{This work was part-funded by the Swiss National Science Foundation (SNF), Project 183569.}
\end{keyword}
\end{frontmatter}



\section{Introduction}
Maintenance of physical equipment, machinery, systems and even complete infrastructure represents an essential process for ensuring successful operation. It helps minimizing downtime of technical equipment \citep{Heidergott.2010}, eliminate the risk thereof \citep{Groenevelt.1992}, or prolong the life of systems \citep{Dogramaci.2004}. Maintenance is often enforced by external factors, such as regulations or quality management \citep{Lee.1987}. Yet maintenance burdens individuals, businesses and organizations with immense costs. For instance, the International Air Transport Association~(IATA) reported that maintenance costs of 49 major airlines increased by over 3~percent from 2012 to 2016, finally totaling \$15.57 billion annually.\footnote{International Air Transport Association~(IATA). \emph{Airline maintenance cost executive commentary}. URL: \url{https://www.iata.org/whatwedo/workgroups/Documents/MCTF/MCTF-FY2016-Report-Public.pdf}, accessed April~18, 2019.}

Decision support in maintenance can be loosely categorized according to two different objectives depending on whether they serve a corrective or preemptive purpose.\footnote{Despite the wealth of earlier works on maintenance operations, there is no universal terminology. Instead, the interested reader is referred to \citet{Jardine.2006}, \citet{Heng.2009}, and \citet{Si.2011} for detailed overviews. We adhere to their terminology.} The former takes place after the failure of machinery with the goal of restoring its operations back to normal. Conversely, preemptive maintenance aims at monitoring these operations, so that the time-to-failure can be predicted and acted upon in order to mitigate potential causes and risk factors by, for instance, replacing deteriorated components in advance. Preemptive actions help in reducing downtime and, in practice, promise substantial financial savings, thus constituting the focus of this paper. 

Preemptive maintenance is based on estimations of the remaining useful life~(RUL) of the machinery. While preventive maintenance makes these forecasts based on human knowledge, predictive maintenance utilizes data-driven models. Different models have been proposed that can be categorized by which input data is utilized (see \Cref{sec:background} for an overview). In the case of raw event data, the conventional approach involves the estimation of probability density functions. If sensor data is available, the prominent approach draws upon machine learning models \citep{Baptista.2018,Seera.2014}. The latter fosters non-linear relationships between sensor observations and RUL estimates, which aid in obtaining more accurate forecasts. 

Machine learning models are subject to an inherent drawback: they frequently operate in a black-box fashion \citep{Breiman.2001,Jang.2019, Subramania.2011}, which, when providing decision support, directly impedes potential insights into the underlying rules behind their decision-making. However, \emph{interpretability} is demanded for a variety of practical reasons. For instance, practitioners desire to benchmark predictive models with their own expertise, as well as to validate the decision-making rules from machine learning models against common knowledge \citep{Delen.2013}. Further, managers can identify potential causes of a short machine lifetime and, thus, outline means by which to reduce errors \citep{Reifsnider.2002}. Moreover, accountability in RUL forecasts is sometimes even required by regulatory agencies, such as, \eg, in aircraft or railroad maintenance \citep[\eg][]{LIU.2006, Papakostas.2010}.

Interpretability refers to machine learning models where the decision logic of the model itself is transparent. Notably, the concept of interpretability differs from post-hoc \emph{explainability} that aims for a different objective. Here, a single (or multiple random) forecast is decomposed, thus highlighting potential relationships but without any structural guarantees \citep{Lipton.2018}. That is, explainability takes an arbitrary model as input and, based on it, attempts to unravel the decision logic behind it, but does so only for a local neighborhood of the input rather than deriving its actual structure. Hence, post-hoc explanations are often not reliable, result in misleading outputs and, because of that, the need for interpretable machine learning has been named an important objective for safety-aware applications \citep{Rudin.2018}. By constructing models that are inherently interpretable, practitioners obtain insights into the underlying mechanisms of the model \citep{Lou.2013}. In keeping with this, we formulate our research objective as follows.

\vspace{0.2cm}
\textsc{Objective:} Forecasting remaining useful life via machine learning with the additional requirement that the model fulfills the definition of \textquote{interpretability}. 
\vspace{0.2cm}

We develop interpretable deep learning models for forecasting RUL as follows: we propose a novel structured-effect neural network that represents a viable trade-off between attaining accurate forecasts and the interpretability from simple distributional estimations. In order to estimate its parameters, we develop an innovative estimation technique based on variational Bayesian inferences that minimize the Kullback-Leibler divergence.\footnote{Some researchers have raised concerns about the applicability of variational Bayesian inference to neural networks, specifically as alternatives might potentially be more straightforward to optimize. Yet variational Bayesian inferences entail obvious strengths in our setting: in contrast to other approaches, it allows to include prior domain knowledge (as is done in our work when choosing regularization priors). }

We demonstrate the effectiveness of our approach in terms of interpretability and prediction performance as follows. We utilize the public \textquote{Turbofan Engine Degradation Simulation} dataset \citep{Saxena.2008} with sensor measurements from aircraft engines. This dataset is widely referred to as a baseline for comparing predictive models in maintenance and RUL predictions; see \eg, \citet{Butcher.2013} and \citet{Dong.2017}. Here the goal is to forecast the remaining useful life until irregular operations, such as breakdowns or failures, take place. The proposed structured-effect neural network outperforms the distribution-based approaches, reducing the forecast error by \SI{51.60}{percent}. While our approach is surpassed slightly by deep learning, it fulfills the definition of being interpretable, \ie, it maintains the same accountability as the much simpler probabilistic approaches. 

The remainder of this paper is structured as follows. \Cref{sec:background} provides an overview on predicting remaining useful life for preemptive maintenance. \Cref{sec:methods} then introduces our methodological framework consisting of probabilistic approaches, machine learning and the novel structured-effect neural network that combines the desirable properties of both. The resulting performance is reported in \Cref{sec:experiments}, where we specifically study the interpretability of the different approaches. Finally, \Cref{sec:discussion} concludes with a discussion of our findings and implications of our work with respect to decision support.

\section{Background}
\label{sec:background}

Previous research has developed an extensive range of mathematical approaches in order to improve maintenance and, due to space constraints, we can only summarize core areas related to our work in the following. For detailed overviews, we refer to \citet{Heng.2009,Liao.2014,Navarro.2010} and \citet{Si.2011}, which provide a schematic categorization of run-to-failure, condition monitoring and predictive methods that estimate the remaining useful life of the machinery. Depending on the underlying approach, the resulting strategy can vary between corrective, responsive, or preemptive maintenance operations. Predictive maintenance, in particular, gives rise to a multitude of variants, \eg, probabilistic approaches and fully data-driven methods that rely upon machine learning together with granular sensor data. The intuition behind inserting sensor measurements into predictive models is that the latter can quantify the environment numerically, the operations and the potential deterioration \citep{Dong.2007}. The observed quantities can be highly versatile and include vibration, oil analysis, temperature, pressure, moisture, humidity, loading, speed, and environmental effects \citep{Si.2011}. As such, sensor measurements are likely to supersede pure condition-based signals in their contribution to overall prognostic capability.

\begin{table}[H]
\scriptsize
\makebox[\textwidth]{%
\begin{tabular}{p{3.5cm}p{2.6cm}p{2.5cm}p{2cm}p{4cm}}
\toprule
\textbf{Approach} &  \textbf{\mcellt{Decision\\ variable}} & \textbf{Input variables} & \textbf{\mcellt{Maintenance\\ strategy}} & \textbf{Operationalization} \\
\midrule
Run-to-failure & --- & --- & Corrective & Service on failure \\[0.5cm]
\midrule
Condition monitoring & Latent state & Event/sensor data & Responsive & Service when latent state indicates (upcoming) failure \\[0.5cm]
Physics-based RUL models & Physics-based RUL & Simulation models & Preemptive & Service when RUL reaches predefined threshold \\[0.5cm]
Probabilistic RUL models & Population-wide (or conditional) RUL & Event data & Preemptive & Service when RUL reaches predefined threshold \\[0.5cm]
\textbf{Sensor-based RUL predictions (\eg, proportional hazards model, machine learning)} &  \textbf{System-specific RUL} & \textbf{Sensor data} & \textbf{Preemptive} & \textbf{Service when RUL reaches predefined threshold} \\[0.5cm]
\bottomrule
\end{tabular}
}
\caption{Schematic overview of key research streams for using RUL models in maintenance operations. Further hybridizations of approaches exists that are not covered by the categorization.}
\label{tbl:maintenance}
\end{table}

In order to carry out preemptive measures, one estimates the \emph{remaining useful life}~(RUL) and then applies a suitable strategy for scheduling maintenance operations (such as a simple threshold rule that triggers a maintenance once RUL undercuts a safety margin) in a cost-efficient manner \citep{Kim.2013, Papakostas.2010, Plitsos.2017}. Mathematically, the RUL at time $t$ can be formalized as a random variable $Y_t$ that depends on the operative environment and its past use $X_t, \ldots, X_2, X_1$, \ie,
\begin{equation}
\expectation \left[Y_t \mid X_t, \ldots, X_2, X_1 \right].
\label{equ:exp_RUL}
\end{equation}
Here the variables $X_1, \ldots, X_t$ can refer to event data tracking past failures \citep{Ghosh.2007}, numerical quantities tracing the machines condition over time as an early warning of malfunctioning \citep{Si.2011}, or measurements of its use as a proxy for deterioration \citep{Navarro.2010}.

\subsection{Probabilistic lifetime models}

Probabilistic models utilize knowledge about the population of machinery by learning from the sensor observations of multiple machines. This knowledge is obtained utilizing predefined probability density functions that specify the probability distributions over machinery lifetimes. Mathematically, when $X_t$ is not available, the RUL estimation turns into $\expectation \left[Y_t \mid X_t, \ldots, X_2, X_1 \right] = \expectation \left[Y_t \right] = \cfrac{\expectation \left[t + Y_t \right]}{R(t)}$, where $R(t)$ is the survival function at $t$. Common choices include exponential, log-logistic, log-normal, gamma, and Weibull distributions \cite[e.\,g.][]{Heng.2009}. We refer to \citet{Navarro.2010} for a detailed survey. For instance, the Weibull distribution has been found to be effective even given few observations of lifetimes, which facilitates its practical use \citep{MAZHAR.2007}. Both log-normal and Weibull distributions can be extended by covariates for sensor-data, which we describe below in \Cref{sec:structured_effect_neural_network} but are then constrained to the mathematical structure, rather than flexibility when calibrating a data-driven approach through machine learning. 

Probabilistic approaches are common choices as they benefit from straightforward use, direct interpretability and reliable estimates, that are often required in practical applications and especially by the regulatory body. However, a focus is almost exclusively placed on raw event data, thereby ignoring the prognostic capacity of sensor data.

Probabilistic approaches can theoretically be extended to accommodate sensor data, resulting in survival models. Since its initial proposal by \Citet{Cox.1972}, the proportional hazards model has been popular for lifetime analysis in general \citep{Wang.2013} and the estimation of RUL in particular. 
 A key advantage of the proportional hazards model over many other approaches is that the interaction between a number of influencing factors can be easily combined with a baseline function that describes the general lifetime of the machinery. More precisely, the proportional hazards model assumes that the probability estimates consists of two components, namely, a structural effect and random effects described by covariates \citep{Si.2011}. As will be discussed later, our structured-effect neural network is built on a similar idea; however, it exploits deep learning to increase the predictive power of the RUL in contrast to the proportional hazards model, which utilizes an exponential model to describe the random effect.

\subsection{Machine learning in lifetime predictions}

Machine learning has recently received great traction for RUL as the flexibility of these models facilitates a superior prognostic capacity. For instance, linear regression models offer the advantage of high interpretability when predicting RUL. Extensions by regularization yield the lasso and ridge regression, which have been found to be effective for high-dimensional sensor data \citep{Zihajehzadeh.2016}. To overcome the limitations of linear relationships, a variety of non-linear models have been utilized, including support vector regression \citep{Baptista.2018}, random forests \citep{Seera.2014}, and neural networks \citep{Riad.2010}. We refer to \citet{Heng.2009} and \citet{Si.2011} for a detailed overview of the proposed models. However, non-linear models generally fall short in terms of their explanatory power \citep{Breiman.2001}. 

Even though machine learning demonstrates high predictive power, these models struggle with the nature of sensor data as time series. It is common practice to make RUL estimates based purely on the sensor data at one specific point in time \citep{Si.2011}. This simplifies $\expectation \left[ Y_t \mid X_t, \ldots X_1 \right]$ to $\expectation \left[ Y_t \mid X_t \right]$, thereby ignoring the past trajectory of sensor measurements. Yet the history of sensor measurements is likely to encode valuable information regarding the past deterioration and usage of machinery. As an intuitive example, a jet engine that experiences considerable vibration might require more frequent check-ups. As a remedy, feature engineering has been proposed in order to aggregate past usage profiles onto feature vectors that are then fed into the machine learning model \citep{Mosallam.2013}. Formally, this yields $\expectation \left[ Y_t \mid \phi(X_t, \ldots, X_1) \right]$ or $\expectation \left[ Y_t \mid X_t, \phi(X_{t-1}, \ldots, X_1) \right]$, where the aggregation function $\phi$ could, for instance, extract the maximum, minimum, or variability from a sensor time series. As a result, the features could theoretically be linked to interpretations but this is largely prohibited by the nature of the machine learning model. 

Advances from deep neural networks have only recently been utilized for the prediction of RUL. In \citet{SateeshBabu.2016}, the authors apply convolutional neural networks along the temporal dimension in order to incorporate automated feature learning from raw sensor signals and predict RUL. In other works, long short-term memory networks~(LSTMs), as a prevalent form of recurrent neural networks, have been shown to perform superior to traditional statistical probability regression methods in predicting RUL \citep{Dong.2017, Wu.2018, Zheng.2017}. Thereby, the LSTM can make use of the complete sequence of sensor measurements by processing complete sequences with the objective of directly estimating the formula $\expectation \left[ Y_t \mid X_t, \ldots, X_1 \right]$ with varying, machine-dependent $t$. In addition, LSTMs entail a high degree of flexibility, which helps to accurately model highly non-linear relationships. This commonly lowers the forecast error, which further translates into improved maintenance operations. 

Deep neural networks are rarely utilized in practical applications for a variety of reasons. Arguably, this is not only because deep neural networks have only recently begun to be used for estimating RUL, but also because they are widely known to be black-box functions with limited to no interpretability. Hence, it is the contribution of this paper to develop a combination of structural predictions and deep learning in order to reach a favorable trade-off between interpretability and prognostic capacity. As points of comparison, we draw upon previous works for RUL predictions, including those concerned with machine learning, feature engineering, and deep learning. 

\subsection{Verification in machine learning}
Interpretability of machine learning is particularly important in mission-critical systems, which requires the development of assessment techniques that reliably identify unlikely types of error \citep{Xiang.2018}. One approach to uncovering cases where the model may be incompatible with the desired behaviour is to systematically search for worst-case results during evaluation \citep[\eg][]{Athalye.2018}. Formal verification proves that machine learning models are specification consistent \citep[\eg][]{Singh.2018}. While the field of formal verification has been subject to research, these approaches are impeded by limited scalability, especially in response to modern deep learning systems.

\subsection{Explainable vs. interpretable machine learning}

Explainable machine learning refers to \emph{post-hoc} explaining predictions without elucidating the mechanisms with which models work. Examples of such post-hoc interpretations are local linear approximation of the model's behavior \citep[\eg partial dependence plots; see][]{Ribeiro.2016} or decompositions of the final prediction into the contribution of each input feature \citep[\eg SHAP values; see][]{Lundberg.2017}. Another widely applied approach to obtaining explanations is to render visualizations to determine qualitatively what a model has learned \citep{vanderMaaten.2008}. However, explainable machine learning is limited in understanding the underlying process of estimation. Notably, it is also limited to a local neighborhood of the input space or the prediction.

In contrast, interpretable machine learning is to encode an interpretable structure \emph{a priori}, which allows to looking into their mechanisms in order to understand the complete functioning of predictions for all possible input features \citep{Murdoch.2019}. Here global relationships are directly encoded in the structure of the model. As such, the relationship in individual features or outcomes for average cases are explicitly modeled. Naturally, linear models have become a prevalent choice for applications in (safety-)critical use cases, where a complete traceability of the model's estimation is inevitable. Hence, the estimation (rather than predictions) can now be compared against prior knowledge or used for obtaining insights.

\section{Methods}
\label{sec:methods}

This research aims at developing forecasting models for the remaining useful life that, on the one hand, obtain a favorable out-of-sample performance while achieving a high degree of interpretability at the same time. Hence, this work contributes to the previous literature by specifically interpreting the relevance of different sensor types and usage profiles in relation to the overall forecast. To date, probabilistic models of failure rates have been widely utilized for predicting the remaining useful life due to their exceptional explanatory power. We thus take the interpretable feature of this approach and develop a method that combines it with the predictive accuracy of deep learning.

We compare the forecasting performance of our structured-effect neural network with the following approaches: (i)~na{\"i}ve empirical estimations, (ii)~probabilistic approaches, (iii)~traditional machine learning, (iv)~traditional machine learning with feature engineering for time series applications, and (v)~deep neural networks. All of the aforementioned methods are outlined in the following.

\subsection{Na{\"i}ve empirical estimation of remaining useful life}
\label{sec:empirical_estimation_of_failure_rates}

Na{\"i}ve empirical estimation of remaining useful life describes the approximation of RUL utilizing past lifetimes of the machinery. Let $Z$ denote the random variable referring to the total lifetime of a machinery, and let $Z_1,\dots,Z_n$ denote $n$ realizations of this random variable. Then we utilize the mean of these realizations to estimate the total lifetime of a machinery, \ie,
\begin{align}
\expectation [ Z ] = \frac{1}{n} \sum_{i = 1}^{n} Z_i.
\end{align}
We can now translate this estimation of the total lifetime into an estimation of the RUL by subtracting the time the machinery has been in use since last being maintained. Let $Y_t$ denote the random variable that describes the RUL of a machinery at time $t$. Then we estimate $Y_t$ by
\begin{align}
\expectation [ Y_t ] = \expectation [ Z ] - t.
\end{align}

\subsection{Probabilistic lifetime models}
\label{sec:pdf}

In accordance with our literature review, we draw upon two prominent probability density functions $P$ that model the lifetime expectancy of machinery, namely, the Weibull distribution, and the log-normal distribution \cite[e.\,g.][]{Vlok.2002}. Let, again, $Z$ denote the random variable referring to the total lifetime of the population. Then the probability density functions of the Weibull distribution and the log-normal distribution at time step $t > 0$ are given by
\begin{align}
P_\text{Weibull}(Z; a, b) &= \frac{b}{a} \left( \frac{Z}{a} \right)^{b - 1} \e{-(\frac{Z}{a})^b}  \quad\text{and} \\
P_\text{log-normal}(Z; a, b) &=  
		\begin{cases}
         	\frac{1}{\sqrt{2 \pi} b Z} \e{-\frac{(log(Z) - a)^2}{2 b^2}}, & Z > 0 , \\
    		0, & Z \leq 0 ,
            \end{cases}
\end{align}
respectively, with distribution parameters $a$ and $b$. All distribution parameters are estimated based on past event data; more precisely, the historical time-spans between failures of the machinery are inserted as the lifetime $Z$. This allows us to estimate the expected lifetime of the machinery after a maintenance event, as well as the corresponding variance.

The mean value of the different probability density functions \emph{could} provide estimates of the remaining useful life for unseen data observations. However, this would ignore the knowledge that the machine has already functioned over $t$ time steps. Hence, we are interested in the conditional expectation, given that the machine had the last maintenance event $t$ time steps ago. This results in an estimated RUL at time $t$ of
\begin{equation}
\label{eqn:RUL_conditional_expectation}
\expectation [ Y_t ] = \expectation_{Z \sim P} \left[ Z \mid Z > t \right] - t .
\end{equation} 
To compute the previous expression, we draw upon the cumulative distribution function $F(Z; \cdot)$ and the definition of the conditional probability. We then rewrite \Cref{eqn:RUL_conditional_expectation} into 
\begin{equation}
\label{eqn:RUL_conditional_expectation2}
\expectation [ Y_t ] = \expectation_{Z \sim P} \left[ Z \mid Z > t \right] - t = \expectation_{Z \sim P} \left[ \frac{Z}{1 - F(Z; \cdot)} \right] - t.
\end{equation} 
Unfortunately, there is no (known) closed-form solution to the expected conditional probability of a Weibull distribution. Hence, we utilize Markov chain Monte Carlo to approximate \Cref{eqn:RUL_conditional_expectation2} for both, the Weibull distribution and the log-normal distribution, in order to come up with the expected remaining lifespan (conditional on the time of the last maintenance event).  

\subsection{Traditional machine learning}
\label{sec:traditional_machine_learning}

In the following, let $f$ refer to the different machine learning models with additional parameters $w$. Then, in each time step $t$, the machine learning model $f$ is fed with the current sensor data $X_t$ and computes the predicted RUL, given by $\tilde{Y}_t = f(X_t; w)$, such that $Y_t \approx \tilde{Y}_t$. The deviation between the true RUL, $Y_t$, and the forecast $\tilde{Y}_t$ defines the prediction error that we try to minimize. Hence, the optimal parameters can be determined by an optimization problem 
\begin{equation}
w^\ast = \argmin_{w}{\norm{ Y_t - f(X_t; w) }} .
\end{equation}

A variety of models $f$ are common in predicting remaining useful life; see the surveys in \citet{Heng.2009} and \citet{Si.2011}. We adhere to previous choices and thus incorporate a variety of baseline models that consist of both linear and non-linear models. Linear models include ridge regression, lasso, and elastic net, all of which are easily interpretable and have been shown to perform well on many machine learning tasks with high-dimensional and even collinear features~\citep{Hastie.2009}. The set of non-linear baseline models include random forest and support vector regression~(SVR).

All models are then fed with two different sets of features: (1)~we take the current sensor measurements $X_t$ when predicting the RUL estimate $Y_t$. However, this approach ignores the trajectory of historic sensor data. (2)~As a remedy, we rely upon feature engineering as a means of condensing the past time series into a feature vector, as described in the following.

Feature engineering provides a means by which to encode the past usage of machinery into an input vector for the predictive model. Yet previous research has only little guidance at hand regarding what type of features are most useful. Hence, we adapt the choice of aggregation functions from \citet{Mosallam.2013}, as detailed in \Cref{tbl:aggregation_functions}. For instance, vibration is known to accelerate deterioration, but it is unclear whether this is caused by sudden peaks (\ie, minima or maxima), frequent changes (\ie, standard deviation), or a constantly high tremor (\ie, average). Mathematically, each aggregation function $\phi$ takes a sequence of past sensor measurements $X_t, \ldots, X_1$ as input and then computes a new input feature $\phi(X_t, \ldots, X_1)$. These aggregation functions are necessary to map the complete trajectory onto a fixed, predefined number of features that can be readily processed by the machine learning models. 

\begin{table}[H]
\centering
\notsotiny
\makebox[\textwidth]{%
\renewcommand{\arraystretch}{1.35}
\begin{tabular}{lll}
\toprule
\textbf{Aggregation function} & \textbf{Formula} & \textbf{Interpretation} \\
\midrule
Max & max($X_1,\dots,X_t$) & Extrema \\
Min & min($X_1,\dots,X_t$) & Extrema \\
Mean & $\mu = \frac{1}{t} \sum_{i = 1}^{t} X_i$ & Average sensor measurement \\
Range & max $-$ min & Variability \\
Sum & $\sum_{i = 1}^t X_i$ & Total signal \\
Energy & $\sum_{i = 1}^{t} X_i^2$ & Total signal with focus on peaks \\
Standard deviation & $\sigma = \sqrt{\frac{1}{t} \sum_{i = 1}^{t} (X_i - \mu})^2$ & Variability \\
Skewness & $\frac{1}{t} \sum_{i = 1}^{t} \left( \frac{X_i - \mu}{\sigma} \right) ^3$ & Symmetry of deviation\\
Kurtosis & $\frac{1}{t} \sum_{i = 1}^{t} \left( \frac{X_i - \mu}{\sigma} \right) ^4$ & Infrequent extreme deviations \\
Peak-to-peak & $\frac{1}{n_1} \sum_{i = 1}^{n_1} \text{loc max} + \frac{1}{n_2} \sum_{i = 1}^{n_2} \text{loc min}$ & Bandwith \\
Root mean square & $\sqrt{\frac{1}{t} \sum_{i = 1}^{t} X_i^2} $ & Total load focus on peaks \\ 
Entropy & $-\sum_{i = 1}^{t} P(X_i) \, \text{log} \, P(X_i) $ & Information signal \\
\lcellt{Arithmetic mean of\\ power spectral density} & $20 \, \text{log}_{10} \cfrac{\frac{1}{t} \sum_{i = 1}^{t} \abs{\text{fft}(X_i)}}{10^{-5}}$ & Frequency of oscillations \\
Line integral & $\sum_{i = 1}^{t - 1} \abs{X_{i + 1} - X_{i}}$ & Path length \\
Kalman filter & $Y_t - b - \sum_{i = 1}^{p} a_i X_{t-i} $ & Unexpected deviation \\
\bottomrule
\end{tabular}
}%
\caption{Our feature engineering draws upon the above aggregation functions. This choice is common in predicting remaining useful life \cite[e.\,g.][]{Mosallam.2013}. Here $p$ refers to an optional parameter specifying the number of lags, which is later set to 50 in accordance with previous research. The expressions loc max and loc min refer to the local maximum and minimum of the inputs.}
\label{tbl:aggregation_functions}
\end{table}

To determine the best hyperparameter combination in traditional machine learning, we implemented group $10$-fold cross-validation in order to minimize the bias associated with random sampling of training and validation data. The group approach also ensures that we do not split maintenance cycles during cross-validation and that the same maintenance cycle is not present in both the training and validation set.

\subsection{Recurrent neural network}
\label{sec:recurrent_neural_network}

\begin{figure}[H]
\centering
\includegraphics[width=.5\textwidth]{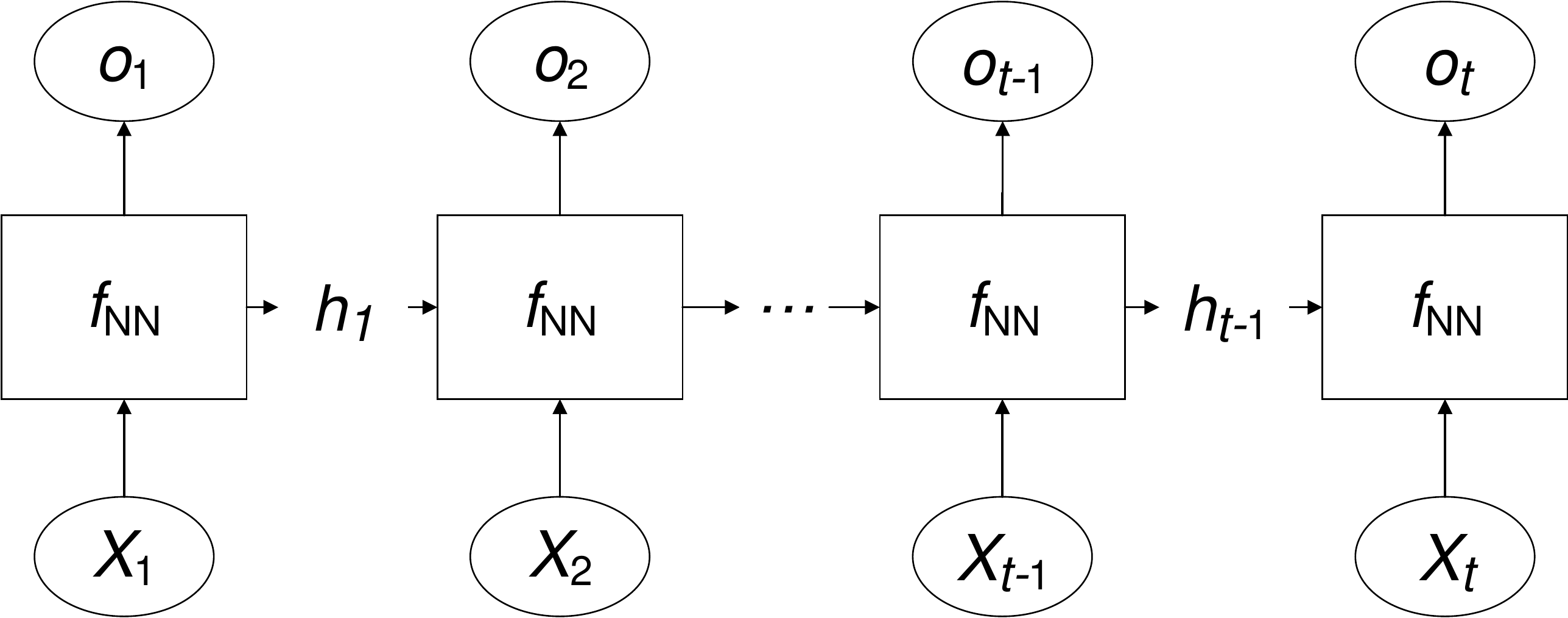}
\caption{Recurrent neural network that recursively applies the same simple neural network $f_\text{NN}$ to the input sequence $X_1, \ldots, X_t$ with outputs $o_1, \ldots, o_t$. The states $h_1, \ldots, h_{t-1}$ encode the previous sequence into a fixed-size feature vector.}
\label{fig:RNN_unrolled}
\end{figure}

Recurrent neural networks refer to a special class of deep neural networks that can learn from sequences of varying lengths, rather than a fixed size of feature vector \citep{Goodfellow.2017}. This is beneficial to our setting, as it allows us to directly inject time series with sensor data into the RNN and predict the remaining useful life from it. The mathematical formalization is as follows: let $f_\text{NN}$ denote a traditional (or deep) neural network that defines a mapping $[X_t , h_{t-1}] \mapsto h_t$ with hidden states $h_{t-1}, h_t \in \mathbb{R}^{n}$ and a suitably chosen dimension $n$. Then a prediction from a complete sequence can be made via 
\begin{equation}
\mathit{RNN}_\Theta = f_\text{NN} ( [X_t, f_\text{NN} ( [X_{t-1}, \ldots f_\text{NN}([X_1, \bm{0}_n]) ] )  ] ) .
\end{equation}
In other words, the RNN iterates over the sequence, while updating its hidden state $h_t$, which summarizes the already-seen sequence, similar to an internal state. This recurrent relationship between the states introduces the possibility of passing information onwards from the current state $h_t$ to the next $h_{t+1}$. Therefore, RNNs can process sequences of arbitrary length, making them capable of utilizing the complete trajectory of sensor data. To illustrate this, \Cref{fig:RNN_unrolled} presents the processing of sequential data by means of unrolling the recurrent structure.

Different variants of recurrent neural networks have been proposed in earlier research; see \citet{Goodfellow.2017}. In this work, we choose the long short-term memory from \citet{Hochreiter.1997} because it is capable to keep information over long sequences, and it enjoys widespread use in research and practical applications \citep{Fischer.2018, Kraus.2018, Srivastava.2018}. For deep neural networks, we reduce the computational runtime for hyperparameter tuning and instead follow conventional guidelines, whereby a random sample of the training data (\SI{10}{\percent}) serves for validation.

\subsection{Proposed structured-effect neural network}
\label{sec:structured_effect_neural_network}

\subsubsection{Model specification}

We now propose our structured-effect model. This approach enforces a specific structure that lends to intuitive interpretation. More precisely, it combines non-parametric approaches for modeling the expected RUL through a probabilistic density function with the flexibility of machine learning in order to incorporate sensor measurements and thus capture the heterogeneity from machine-specific deterioration processes.

The idea of building structured models loosely resembles earlier research efforts related to proportional hazards models \citep{Cox.1972} that present a popular choice in lifetime analysis. This class of survival models decomposes its estimations into components referring to the baseline function and a function of covariates: the baseline specifies the general lifetime across all machines via the same function $\lambda(t)$. The latter further assumes machine-specific effects through additional covariates that describe the random effects. Our structured-effect neural network follows a similar intuition, as it assumes a population-wide general lifetime common across all machines and further sensor-based deviations in order to model the within-machine heterogeneity due to the different usage profiles. 

Our structured-effect model splits the estimated remaining useful life into three components, namely, a non-parametric baseline, a covariate-based prediction, and a recurrent component which specifically incorporates the historic trajectory of sensor measurements. These components help in explaining the variance among the different machine lifetimes and, for this purpose, we again draw upon the history of sensor measurements $X_t, \ldots, X_1$. Let $\lambda(t)$ denote the non-parametric part with the explicit probabilistic lifetime model and let further $\mathit{RNN}_\Theta$ refer to a recurrent neural network (such as a long short-term memory) with weights $\Theta$. Then the prediction of the structured-effect neural network $\mathit{SENN}_\Theta(t; X_t, \ldots, X_1)$ follows the form 
\begin{equation}
\label{eq:structured_effect_neural_network}
\footnotesize
\mathit{SENN}_\Theta(t; X_t, \ldots, X_1)  = \underbrace{\lambda(t)}_{\text{\mcellt{Non-parametric component\\[-0.6em] with explicit lifetime model}}} + \underbrace{\beta^T X_t}_{\text{\mcellt{Linear component\\[-0.6em] with current condition}}} + \underbrace{\mathit{RNN}_\Theta(X_t, \ldots, X_1, t)}_{\text{\mcellt{Recurrent component\\[-0.6em] with deep neural network}}} 
\end{equation} 
with coefficients $\beta$. Model variations are discussed later in \Cref{sec:model_variations}.

While our model follows a similar intuition behind the proportional hazards model in decomposing the prediction, it also reveals clear differences, as it introduces a recurrent neural network that allows for considerably higher flexibility in modeling the variance and even incorporates the \emph{complete} sequence of sensor measurements and not just a simple vector of covariates. Moreover, the specific way of our model formulation entails a set of further advantages. On the one hand, it circumvents again the explicit need for feature engineering. On the other hand, it achieves a beneficial trade-off between interpretability of non-parametric approaches and the flexibility of non-linear predictions from sensor data. Here the deep neural network needs to explain a considerably smaller variance compared to an approach based solely on a neural network, thereby facilitating the estimation of the network weights. To this end, practitioners can decompose the prediction into a population-wide baseline and machine-specific heterogeneity, based on which they can explicitly quantify the relative contribution of each components through the corresponding coefficients. As such, one can identify reasons why the remaining useful life attains a certain value (\eg, a negative value from the recurrent component indicates a strong deterioration over time) or one can attribute deterioration to unexpected behavior. Moreover, the proposed approach is highly extensible and can easily be generalized to other parameterizations or domains. 

We later experiment with different variations of the structured-effect model. These differ in the choices with which we specify the different components. First, we adhere to conventional approaches in predictive maintenance \citep{Navarro.2010} by assuming that the lifetimes follow either a conditional Weibull or a conditional log-normal distribution. That is, we obtain 
\begin{equation}
\lambda(t) = \expectation_{Z \sim \text{Weibull}(a, b)} [ Z \mid Z > t ] - t 
\quad\text{and}\quad
\lambda(t) = \expectation_{Z \sim \text{log-normal}(a, b)} [ Z \mid Z > t ] - t 
\end{equation}
with distribution parameters $a$ and $b$. Thus, the first component is identical to the probabilistic lifetime models that we utilize as part of our benchmarks. Second, the linear component can either be fed directly with $X_t$ or, alternatively, one could also apply feature engineering to it, \ie, giving $\phi(X_t, \ldots, X_1)$. The benefit of the latter is that we again obtain a linear structure where one can assess the relevance of individual predictors by looking at the coefficients. Here we further assume a linear combination as used in ordinary least squares and, as an extension, introduce priors, so that we yield a regularization, where the coefficients in the linear component are estimated via the least absolute shrinkage operator~(lasso). This performs implicitly variable selection in the linear component as some coefficients are directly set to zero \citep{Tibshirani.1996}. Third, the recurrent neural network is implemented via a long short-term memory as this represents the state-of-the-art in sequence learning \citep{Goodfellow.2017}.

\subsubsection{Model estimation through variational Bayesian inferences}

We now detail how we estimate the parameters inside the structured-effect neural network. We refer to $\theta$ as the combined set of unknown parameters and $X$ as the overall dataset including all sensor measurements. Then the objective is to determine the optimal parameters
\begin{equation}
\label{eqn:theta_star}
\theta^\ast = \argmax_{\theta} \; P(\theta \,|\, X) .
\end{equation}
We solve the previous optimization problem through a variational Bayesian method. The predominant reason for this choice over traditional optimization is that the latter would merely give point estimates of the different parameters, whereas variational Bayesian inferences yield quantifications of uncertainty. For instance, this allows us to obtain confidence assessments concerning the relative importance of the different components and thus facilitates the interpretability of our approach. 

In our model estimation, we treat all parameters as latent variables with a pre-defined prior distribution and, subsequently, maximize the overall likelihood of the parameters according to the following procedure. That is, utilizing  Bayes' theorem, \Cref{eqn:theta_star} is rewritten to
\begin{equation}
P(\theta \,|\, X) = \frac{P(X \,|\, \theta)\,P(\theta)}{P(X)} 
= \frac{P(X \,|\, \theta)\,P(\theta)}{\int \! P(X \,|\, \theta)\, P(\theta) \, \text{d}\theta} .
\end{equation} 
As a result, the denominator can be computed through sampling methods, with the most prominent being Markov chain Monte Carlo~(MCMC). However, MCMC methods are computationally expensive as the runtime scales exponentially with the dimensions of $\theta$. Thus, this algorithm becomes intractable for large-scale or high-dimensional datasets. As a remedy, we propose the use of variational Bayes for approximating the posterior distributions. We derive a variational lower bound, called ELBO, for our structured-effect neural network in \Cref{sec:derivation}. 

\subsubsection{Estimation parameters}

In our experiments, we optimize the $\mathit{SENN}$-model by utilizing the Adam optimizer with learning rate \num{0.005} and all other parameters set to the default values. All implementations are performed in Python utilizing the probabilistic programming library \textquote{pyro} (\url{http://pyro.ai/}). Code for reproducibility is available online.\footnote{See \url{https://github.com/MathiasKraus/PredictiveMaintenance}}.

As part of our computational experiments, we later draw upon the following architectures of the structured-effect neural network: (1)~we assume the non-parametric component to follow a Weibull or log-normal prior distribution, where the underlying distribution parameters are modeled as informative normal prior distributions. Mathematically, this is given by $a \sim \mathcal{N}(a_{\text{empirical}},\,1)$ and $b \sim \mathcal{N}(b_{\text{empirical}},\,1)$. (2)~The linear component is modeled such that the coefficients stem from normal prior distributions (\ie, as used in ordinary least squares). This is formalized by $\beta_i \sim \mathcal{N}(0,\,10)$, where we allow for a wider standard deviation to better handle variations in the relative influence of the predictors. As an alternative, we also implement weakly informative prior distributions (\ie, Laplace priors). The latter enforce a regularization similar to the least absolute shrinkage operator in the sense that certain coefficients are set exactly to zero in order to perform implicit variable selection and come up with a parsimonious model structure. (3)~The recurrent component is implemented as a long short-term memory network with two layers containing \num{100} and \num{50} neurons, respectively. To reduce computational costs, we follow common approaches and utilize the trajectory of the previous \num{50} sensor values at all time steps. All weights in the network are implemented as variational parameters that follow a Gaussian prior with standard deviation of 1. Utilizing \Cref{equ:grad}, we optimize the three components simultaneously.

\section{Computational experiments}
\label{sec:experiments}

\subsection{Dataset}

For reasons of comparability, all computational experiments are based on the \textquote{Turbofan Engine Degradation Simulation} dataset, which is widely utilized as a baseline for comparing predictive models in maintenance and RUL predictions; see \eg, \citet{Butcher.2013} and \citet{Dong.2017}. The objective is to predict the RUL (measured in cycles) based on sensor data from 200 aircraft engines.\footnote{The specific dataset of this study can be downloaded from \url{https://ti.arc.nasa.gov/tech/dash/groups/pcoe/prognostic-data-repository/}, accessed April~18, 2019.} More specifically, it includes measurements from 21 sensors. Unfortunately, however, the exact name of each sensor is sanitized. In addition, the dataset comes with a pre-determined split into a training set (100 engines) and a test set (also 100 engines). The average RUL spans \num{82.30} cycles with a standard deviation of \num{54.59}. Moreover, half of the engines experience a failure within \num{77} cycles, while only \SI{25}{\percent} exceed \num{118} cycles.

\subsection{Prediction performance for remaining useful life}

The prediction results for all models are listed in \Cref{tbl:results}. Here we report the mean absolute error, as it represents a widely utilized metric for this dataset \citep{Ramasso.2014}. The benefit of this metric is that practitioners can easily translate the forecast error into a number of cycles that would serve as a security margin. The table also compares two different feature sets for traditional machine learning, \ie, on which we use only the sensor measurements from the current time step or on which we additionally apply aggregation functions to the sensors as part of feature engineering. 

The empirical RUL in the first row reflects the performance of our na{\"i}ve benchmark when using no predictor (\ie, predicting the average RUL of the machines). The following conditional expectations are based on the Weibull and log-normal distribution, that result in improvements of \SI{39.17}{\percent} and \SI{38.32}{\percent}, respectively. Among the traditional machine learning models, we find the lowest mean absolute error when using the random forest, which yields an improvement of \SI{35.08}{\percent} compared to the log-normal-based conditional expectation. Thereby, our results identify a superior performance through the use of feature engineering for the majority of traditional machine learning models. Recurrent neural networks outperform traditional machine learning. In particular, the LSTM yields the overall lowest mean absolute error, outperforming the random forest with feature engineering by \SI{37.12}{\percent} and the empirical RUL by \SI{60.51}{\percent}. 

The structured-effect neural networks outperform traditional machine learning. Utilizing a log-normal distribution along with feature engineering yields an improvement of \SI{25.44}{\percent} compared to the best traditional machine learning model. Thereby, feature engineering accounts for \SI{11.91}{\percent} of the improvement, strengthening the assumption that feature engineering of sensor data facilitates the prediction of RUL. 

\begin{table}[H]
\tiny
\makebox[\textwidth]{%
\begin{tabular}{ll S *{4}{S[table-align-text-post=false]} *{4}{S[table-align-text-post=false]}}
\toprule
\multicolumn{2}{l}{\textbf{Method}} & 
\multicolumn{1}{c}{\textbf{MAE}} & \multicolumn{4}{c}{\textbf{Forecast comparison ($\bm{t}$-statistic)}} & \multicolumn{4}{c}{\textbf{Forecast comparison ($\bm{P}$-value)}}  \\
\cmidrule(l){3-3} \cmidrule(l){4-7} \cmidrule(l){8-11}
\multicolumn{2}{l}{} & & \multicolumn{1}{c}{Best} & \multicolumn{1}{c}{Best} & \multicolumn{1}{c}{Best} & \multicolumn{1}{c}{Best} & \multicolumn{1}{c}{Best} & \multicolumn{1}{c}{Best} & \multicolumn{1}{c}{Best} & \multicolumn{1}{c}{Best} \\
\multicolumn{2}{l}{} & & \multicolumn{1}{c}{baseline} & \multicolumn{1}{c}{machine} & \multicolumn{1}{c}{LSTM} & \multicolumn{1}{c}{structured-effect} & \multicolumn{1}{c}{baseline} & \multicolumn{1}{c}{machine} & \multicolumn{1}{c}{LSTM} & \multicolumn{1}{c}{structured-effect} \\
\multicolumn{2}{l}{} & & & \multicolumn{1}{c}{learning} & & \multicolumn{1}{c}{LSTM} & & \multicolumn{1}{c}{learning} & & \multicolumn{1}{c}{LSTM} \\
\midrule
\multicolumn{6}{l}{\textsc{Baselines without sensor data}} \\
\multicolumn{2}{l}{\quad Empirical RUL} & 45.060 & 8.286 & 24.933 & 16.974 & 18.865 & 0.468 & 0.486 & 0.778 & 0.642 \\  
\multicolumn{2}{l}{\quad Conditional expectation (Weibull)} & 27.794 & 0.288 & 8.309 & 9.615 & 8.018 & 0.293 & 0.462 & 0.666 & 0.457 \\ 
\multicolumn{2}{l}{\quad \textbf{Conditional expectation (log-normal)}} & \bfseries 27.409 & {---} & 4.420 & 15.269 & 10.285 & {---} & \bfseries 0.464 & \bfseries 0.669 & \bfseries 0.451 \\
\midrule
\multicolumn{6}{l}{\textsc{Traditional machine learning}} \\
\multicolumn{2}{l}{\quad Ridge regression} & 19.193 & -6.789 & 1.438 & 5.270 & 2.768 & 0.012* & 0.139 & 0.450 & 0.322 \\
\multicolumn{2}{l}{\quad Ridge regression (with feature engineering)} & 18.382 & -8.029 & 0.427 & 3.297 & 2.778 & 0.010* & 0.132 & 0.343 & 0.399 \\[0.25em]
\multicolumn{2}{l}{\quad Lasso} & 19.229 & -7.015 & 0.766 & 6.577 & 4.145 & 0.015* & 0.222 & 0.500 & 0.401 \\
\multicolumn{2}{l}{\quad Lasso (with feature engineering)} & 18.853 & -7.842 & 0.550 & 5.949 & 2.324 & 0.014* & 0.293 & 0.432 & 0.390 \\[0.25em]
\multicolumn{2}{l}{\quad Elastic net} & 19.229 & -7.276 & 0.990 & 5.190 & 2.829 & 0.015* & 0.222 & 0.500 & 0.401 \\
\multicolumn{2}{l}{\quad Elastic net (with feature engineering)} & 18.245 & -9.055 & 0.458 & 4.244 & 2.572 & 0.009** & 0.132 & 0.297 & 0.245 \\[0.25em]
\multicolumn{2}{l}{\quad Random forest} & 17.884 & -4.927 & 0.058 & 5.909 & 4.487 & 0.006** & 0.102 & 0.240 & 0.198 \\
\multicolumn{2}{l}{\quad \textbf{Random forest (with feature engineering)}} & \bfseries 17.793 & -9.495 & {---} & 2.924 & 3.263 & \bfseries 0.006** & {---} & \bfseries 0.236 & \bfseries 0.180 \\[0.25em]  
\multicolumn{2}{l}{\quad SVR} & 18.109 & -7.321 & 0.240 & 5.756 & 3.976 & 0.011* & 0.129 & 0.288 & 0.230 \\ 
\multicolumn{2}{l}{\quad SVR (with feature engineering)} & 21.932 & -4.706 & 3.081 & 9.440 & 3.740 & 0.092* & 0.310 & 0.583 & 0.531 \\ 
\midrule
\multicolumn{6}{l}{\textsc{Recurrent neural networks}} \\
\quad \textbf{LSTM} & & \bfseries 11.188 & -11.596 & -4.981 & {---} & -1.441 & \bfseries 0.000*** & \bfseries 0.000*** & {---} & \bfseries 0.003** \\
\midrule
\multicolumn{6}{l}{\textsc{Structured-effect neural networks}} \\
\quad \emph{Distribution} & \emph{Linear component} \\
\cmidrule(l){1-1}\cmidrule(l){2-2}
\quad Weibull & None & 15.862 & -11.266 & -1.015 & 4.424 & 1.825 & 0.000*** & 0.066* & 0.255 & 0.200 \\
\quad Weibull & Regularized & 17.433 & -8.617 & -0.213 & 3.536 & 2.746 & 0.000*** & 0.094* & 0.261 & 0.220 \\
\quad Weibull & Feature engineering & 13.392 & -8.526 & -2.595 & 1.352 & 0.068 & 0.000*** & 0.004** & 0.144 & 0.134 \\
\quad Weibull & Regularized feature engineering & 14.989 & -10.918 & -1.579 & 1.710 & 1.381 & 0.000*** & 0.049* & 0.284 & 0.183 \\[0.25em]
\quad log-normal & None & 15.061 & -6.294 & -1.420 & 1.627 & 1.702 & 0.000*** & 0.057* & 0.261 & 0.198 \\ 
\quad log-normal & Regularized & 16.319 & -8.200 & -0.779 & 4.582 & 1.334 & 0.000*** & 0.094* & 0.310 & 0.211 \\
\quad \textbf{log-normal} & \textbf{Feature engineering} & \bfseries 13.267 & -13.620 & -2.912 & 1.764 & {---} & \bfseries 0.000*** & \bfseries 0.000*** & \bfseries 0.142 & {---} \\
\quad log-normal & Regularized feature engineering & 14.545 & -11.331 & -1.736 & 3.293 & 0.651 & 0.000*** & 0.038* & 0.252 & 0.162 \\[0.25em]
\bottomrule
\multicolumn{4}{l}{\textbf{Significance level: * 0.1, ** 0.01, *** 0.001}}
\end{tabular}
}%
\caption{Comparison of prediction performance over remaining useful life across different model specifications. Here we specifically report whether the models only utilize sensor measurements from the current time step or whether aggregation functions have been applied to it as part of feature engineering. Consistent with earlier works \citep{Ramasso.2014}, the mean absolute error (MAE) is given. The best-performing model in each panel is highlighted in bold. Additionally, we perform $t$-tests between each model and the best performing model from each of the four categories. The $t$-tests are based on the MAE of the forecasted RUL to show that improvements are at a statistically significant level.}
\label{tbl:results}
\end{table}

\subsection{Forecast decomposition for RUL predictions}

We now demonstrate how the proposed structured-effect model achieves accountability over its RUL forecasts. That is, we leverage the linear model specification and compute the estimated values for each summand in \Cref{eq:structured_effect_neural_network} when making a RUL prediction. Yet the model can still adapt to non-linearity since the neural network can absorb the variance that cannot be explained by the other components. 

\begin{figure}
\vspace{-1cm}
\centering
\makebox[0.7\textwidth]{%
\includegraphics[width=0.7\linewidth]{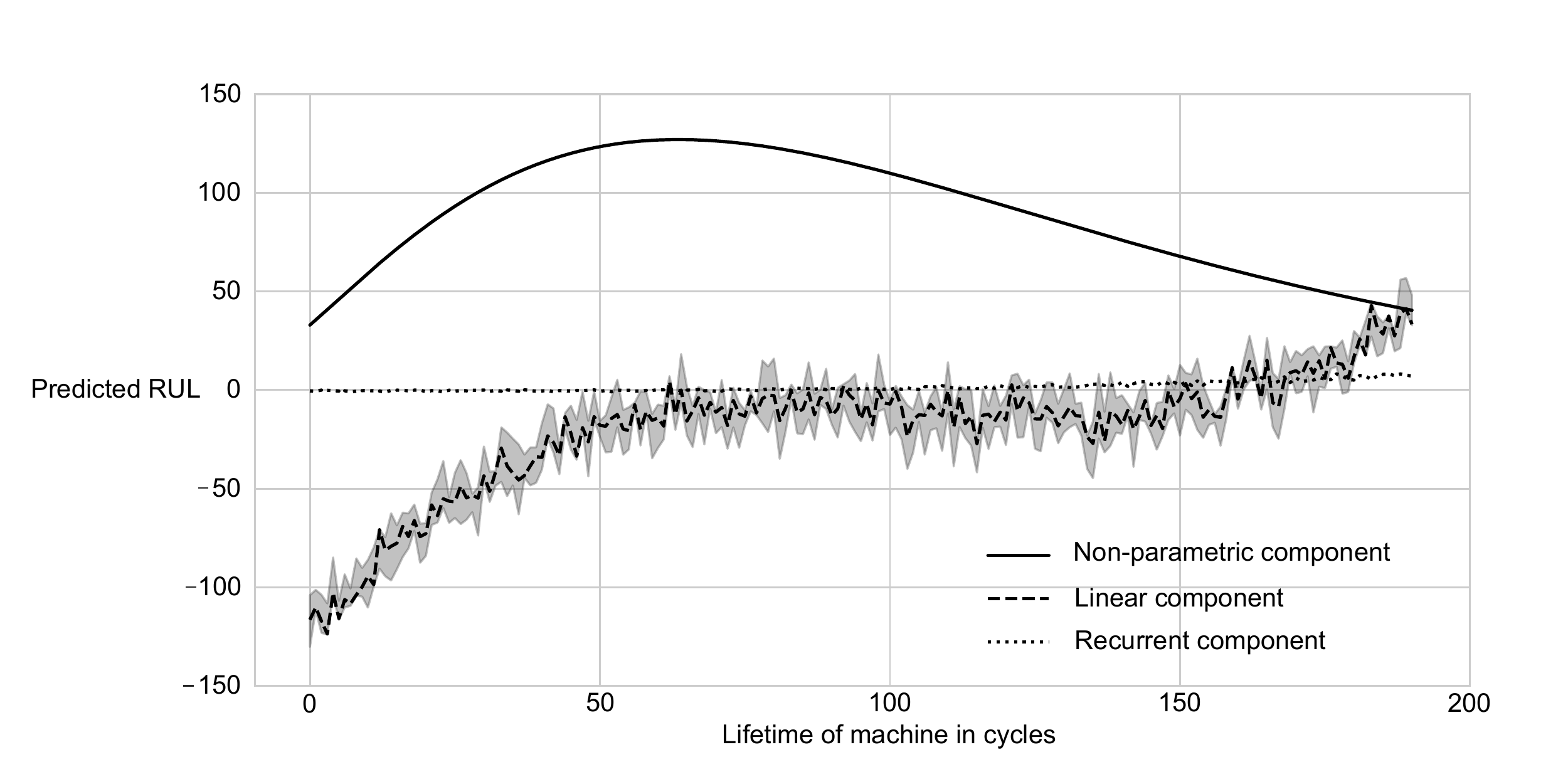}
}%
\caption{This plot visualizes the RUL predictions made by the structured-effect LSTMs based on a log-normal and normal priors in the linear component for an example engine. It decomposes the forecasts using the structured-effect model into three components that facilitate interpretations of how predictions are formed. (1)~The distribution-based lifetime component contributes a considerable portion of the overall forecasts, as it is well suited to model the overall nature of the remaining useful life. This part is identical across all engines. (2)~The sensor measurements introduce a variability that adapts to the specific usage profile of an engine. This component originates from a Bayesian linear model and we can trace the forecast back to individual sensors. (3)~The recurrent neural network introduces a non-linear component that operates in black-box fashion.}
\label{fig:decomposition}
\end{figure}

\Cref{fig:decomposition} illustrates the interpretability of the RUL forecasts for an example engine. More specifically, we can understand how predictions are formed by decomposing the forecasts from the structured-effect model into three components -- namely, the probabilistic RUL model, the linear combination of sensor measurements, and an additional neural network -- as follows:
\begin{enumerate}
\item As we can see, the distribution-based lifetime component accounts for a considerable portion of the forecast. The maximum values in the example exceed \num{100}, which is considerably higher than the maximum value computed by the recurrent component. The component reaches this value when making a prediction after around \num{50} cycles and, with each subsequent usage cycle, lowers the estimated remaining useful life.  
 Notably, it is identical across all engines as it encodes the prior knowledge before considering the engine-specific deterioration process. 
\item The second component specifies a linear combination of sensor measurements, which allows the predictions to adapt to the specific usage profiles of individual engines and explains the within-engine and within-time variability. It thus no longer yields a smooth curve but rather an engine-specific pattern. Formally, this component refers to $\beta^T X_t$ and, in order to determine the relevance of sensor $i$, we simply interpret the coefficients in the vector $\beta$. 
\item While the previous linear component still achieves full accountability over its forecasts, we now introduce the final component for modeling the remaining noise. Here we draw upon (deep) neural networks, as they are known to effectively model non-linearities. However, we thus lose the explanatory power for this component, as neural networks largely operate in a black-box fashion. In our example, we see that the recurrent part entails a non-linear curve but takes higher values in later cycles. This indicates that a linear combination is not always sufficient for making predictions and, as a remedy, the structured-effect model can benefit from additional non-linear relationships and from accumulating the usage profile over time.

Notably, the magnitude of the recurrent component is much smaller than the magnitude of the other components. This is beneficial, as the SENN attributes most of the explained variance to other, interpretable model components. Methodologically, it is likely to be based on the following: at timestep $t$, the SENN makes prediction of the RUL from the current sensor data $X_t$, and the trajectory of sensor data $X_t, X_{t-1}, X_{t-2},\dots, X_1$. As shown in \Cref{tbl:results}, $X_t$  is highly informative for estimating RUL and, by following stochastic gradient descent, it optimizes in the direction where the loss function decreases the most (\ie in the direction of both the distribution-based and the linear component). Only after optimizing the the interpretable components, the model updates the recurrent component to further push predictive performance via non-linear mappings.

\end{enumerate}

We can further compute the fraction of variance explained by the different components relative to the overall variance of the actual RUL values. Thereby, we quantify the contribution of each component to the overall forecast. Accordingly, this is defined by 
\begin{equation}
1 - \frac{ \sum_t \left( \tilde{Y}_t - \tilde{\Psi}_t \right)^2  }{\sum_t \left( \tilde{Y}_t - \frac{1}{T} \sum_t \tilde{Y}_t \right)^2} ,
\end{equation}
where $T$ denotes the total number of observations and $\tilde{\Psi}_t$ is the prediction of the component under study. Accordingly, we obtain a score of \num{0.175} for the non-parametric part, \num{0.408} for the linear component, and \num{0.064} for the recurrent component. This matches our expectations and, once more, highlights the overall importance of the distribution-based part of the overall forecast, as well as the role of the neural network in modeling secondary variations.

\subsection{Estimated parameters}

\begin{figure}
\centering
\makebox[\textwidth]{%
\includegraphics[width=0.7\linewidth]{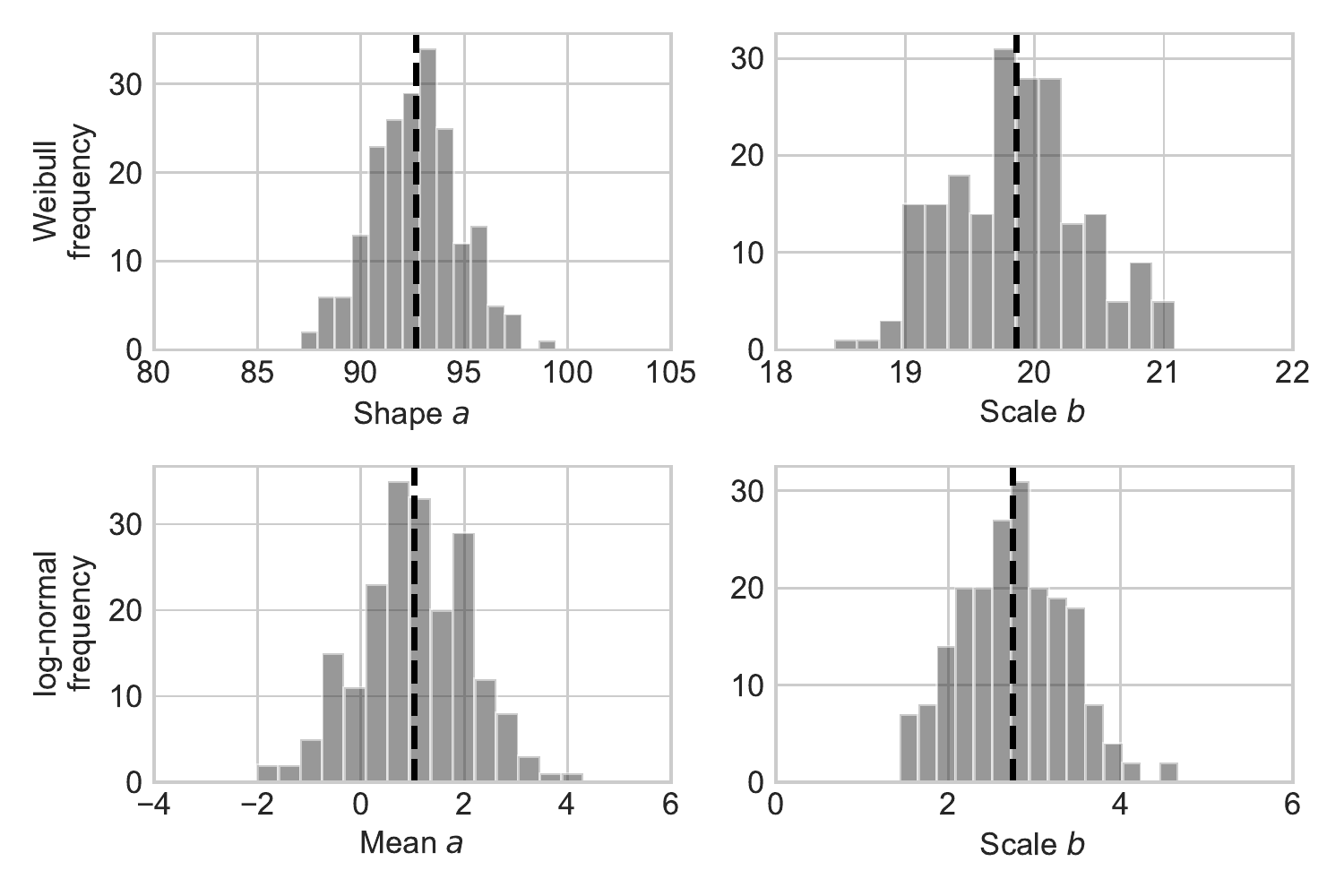}
}%
\caption{These histograms illustrate the posterior distribution of the estimated parameters (\ie, shape, mean and scale) inside the structured-effect neural network (with log-normal structure and normal priors in the linear component). Here the estimations are compared for both distributions, namely, the Weibull and log-normal distributions. Altogether, the posteriors quantify the uncertainty of the estimated parameters.}
\label{fig:posterior_dist}
\end{figure}

It is common practice to compute parameters in predictive models as point estimates \citep{Hastie.2009}, while our optimization technique based on variational Bayesian inferences allows for uncertainty estimates. This yields a key benefit, since we can validate our confidence in the structural components of the model by studying the posterior distribution of the parameter estimates. \Cref{fig:posterior_dist} depicts these parameters specifying the Weibull and log-normal distribution inside the structured-effect models. 

\Cref{tbl:posterior_linear_component} further reports the posterior distribution of the coefficients $\beta_i$ from the linear component of the structured-effect neural network (shown is the SENN model based on a log-normal and normal priors for $\beta_i$). These measure the effect size, \ie, how a change in a sensor measurement affects the forecast. We see that the confidence regions for the different coefficients vary considerably. For reasons of comparability, we further report the standardized coefficients $\beta_i \, \text{var}({\beta_i}) / \text{var}(Y_1, \ldots, Y_t)$, which correct for the variance of the predictor, as well as the outcome \citep{Bring.1994}. As a result, this value allows us to rank variables by their importance. 

\begin{table}[H]
\notsotiny
\makebox[\textwidth]{%
\centering
\begin{tabular}{l SSS}
\toprule
\textbf{Sensor} & \textbf{Mean} & \textbf{Standard} & \textbf{Standardized} \\
& \textbf{estimate} & \textbf{deviation} & \textbf{coefficient} \\
\midrule
$X_9$ & -33.169 & 0.498 & -16.506 \\
$X_{12}$ & 49.721 & 0.250 & 12.440 \\
$X_{21}$ & 44.932 & 0.258 & 11.601 \\
$X_7$ & 48.154 & 0.230 & 11.073 \\
$X_{11}$ & -24.622 & 0.357 & -8.796 \\
$X_{20}$ & 42.850 & 0.184 & 7.880 \\
$X_{14}$ & -22.540 & 0.317 & -7.155 \\
$X_4$ & -16.183 & 0.275 & -4.447 \\
$X_{15}$ & -13.934 & 0.297 & -4.145 \\
$X_2$ & -12.446 & 0.316 & -3.931 \\
$X_6$ & 19.678 & 0.159 & 3.126 \\
$X_3$ & -7.851 & 0.318 & -2.494 \\
$X_{17}$ & -9.951 & 0.208 & -2.066 \\
$X_8$ & -4.121 & 0.367 & -1.511 \\
$X_{16}$ & -0.273 & 2.105 & -0.574 \\
$X_{13}$ & -1.424 & 0.348 & -0.495 \\
$X_{19}$ & 0.234 & 2.028 & 0.474 \\
$X_{10}$ & 0.126 & 1.900 & 0.239 \\
$X_{18}$ & -0.095 & 1.936 & -0.184 \\
$X_1$ & -0.070 & 1.937 & -0.135 \\
$X_5$ & 0.001 & 1.993 & 0.002 \\
\bottomrule
\end{tabular}
}%
\caption{Reported here are the posterior estimates of the effect size as measured by the coefficients $\beta_i$ inside the linear component of the structured-effect neural network. The coefficients entail direct interpretations (similar to ordinary least squares) as to how a certain percentage of change in a sensor measurement affects the RUL prediction. In addition, standardized coefficients are reported, as they allow for the ranking of variables by importance.}
\label{tbl:posterior_linear_component}
\end{table} 

\subsection{Model variations}
\label{sec:model_variations}

We experimented with alternative specifications of our structured-effect neural network as follows. 

First, we extended the neural network by an additional weighting factor $\gamma$. This yields a component $\gamma \, \mathit{RNN}_\Theta$. However, it resulted in an inferior performance in all of our experiments due to severe overfitting. 

Second, we experimented with a two-stage estimation approach. Here we first optimized the non-parametric component and the linear component by traditional gradient descent. Afterwards, we optimized the recurrent component against the residuals from the first stage. This approach is generally easier to train as there are fewer parameters in each stage. Yet we found an inferior performance as compared to the proposed $\mathit{SENN}$: the mean absolute error increased to \SI{16.209}. This is possibly owed to the fact that it prevents information sharing between the different components. Details are reported in \Cref{sec:two_stage_optimization}.

\section{Discussion}
\label{sec:discussion}

\subsection{Implication for decision support}

Estimates of the remaining useful life can facilitate decision support with the objective of replacing deteriorated components and thus mitigating potential risks and failures that generally result in increased costs. 
 Our approach to predicting remaining useful life is thus of direct relevance to practitioners. According to a McKinsey report, the use of accurate prediction models for RUL as a cornerstone for predictive maintenance can typically reduce the downtime of machinery by \SIrange{30}{50}{\percent} and, at the same time, increase the overall life of machines by \SIrange{20}{40}{\percent}.\footnote{McKinsey (2017). Manufacturing: Analytics unleashes productivity and profitability. URL: \url{https://www.mckinsey.com/business-functions/operations/our-insights/manufacturing-analytics-unleashes-productivity-and-profitability}, last accessed on April~18, 2019.} By knowing the exact time-to-failure, companies can plan maintenance ahead of time and, therefore, make preparations for efficient decision support. As a result, even small improvements in predictive power translate into substantial operational cost savings. 

As a direct implication for management, this research shows that forward-looking predictive analytics is capable of heavily influencing the way decision support in maintenance operations is conducted. However, predictive models are most powerful when fed by a large number of predictors (\ie sensors) that describe the condition of the machinery and the environmental effects that influence the system. Therefore, managers should encourage the implementation of additional sensors to further improve accuracy when forecasting the time-to-failure. Moreover, investments in artificial intelligence are oftentimes necessary for the majority of firms who have not yet taken their first step into the age of deep learning. 

\subsection{Implications for the use of analytics}

Trajectories of sensor data accumulate relevant information regarding the past usage profile of the machinery and, thereby, facilitate a prognosis regarding the risk of failure. Mathematically, this results in the objective of finding a mapping $f : \left[ X_1, \ldots, X_t \right] \mapsto Y_t$ that is not dependent on the current time step $t$ and thus utilizes a time series with historic measurements of arbitrary length in order to infer a prediction from it. The task can be accomplished by a special type of deep learning -- namely, recurrent neural networks -- as these networks can sequentially process past measurements and store the processed knowledge in their hidden layers. Even though the benefits are obvious, the use of such networks in decision support systems research remains scarce with few exceptions \citep[\eg][]{Evermann.2017, Kraus.2017, Mahmoudi.2018}.

Deep learning is often believed to require extensive amounts of data in order to be successful. However, our approach, based on variational Bayesian estimation, represents a viable alternative that can overcome this limitation. Instead of vast quantities of input data, it advocates domain knowledge that is explicitly encoded in a structural model. In our case, we already know the approximate shape of the predicted variable and can incorporate this via a probability density function into our structural part of the model. As a result, the predetermined structure can be fitted fairly easily with variational inference and thus presents a path towards encoding domain knowledge into deep neural networks. The structured effect reduces the variance and thus makes it easier to describe the remaining variance with a neural network. 

\subsection{Implications from interpretable forecasts}

Our approach contributes to interpretability of deep learning. Here we remind the reader of the difference between explainability and interpretability in machine learning \citep{Lipton.2018,Lou.2013,Rudin.2018}. Explainability merely allows a post-hoc analysis of how predictions were computed in a local neighborhood. In contrast, interpretability presents a stronger notion: it requires machine learning models to attain complete transparency of their decision logic. Thus, we contribute to a novel approach for interpretable machine learning to decision support, that can eventually benefit (safety-)critical application fields where accountable models are required.   

The high degree of interpretability of our approach reveals further implications. In practice, gaining insights into the estimated RUL aids engineers in identifying potential risks and weak spots when designing machinery. For instance, a high coefficient for a sensor measuring moisture could encourage designers to improve the sealing of a given piece of machinery. By shedding light on the prediction process, structured-effect neural networks enable novel conclusions regarding the relevance of each sensor. 

Decision support as a discipline takes the demands of all stakeholders into account. With regard to the latter, managers, for instance, need to understand the decision-making of automated systems. However, this requirement is not fulfilled by recent trends in advanced analytics and especially deep learning, as these mostly operate in a black-box fashion \cite[e.\,g.][]{Goodfellow.2017}. As a remedy, our structured-effect neural network shows improvements in predictive performance as compared to traditional machine learning, while also allowing for a high degree of interpretability. \Cref{tbl:PredVsInterp} compares the stylized characteristics of our structured-effect neural network to other approaches.

\begin{table}[H]
\scriptsize
\makebox[\textwidth]{%
\begin{tabular}{p{2cm} p{4cm}p{4.5cm}p{4cm}p{3.5cm}}
\toprule
\textbf{Method} & \textbf{Predictive performance} & \textbf{Interpretability} & \textbf{Non-linearities} & \textbf{Estimation}\\
\midrule
Probabilistic models & Poor (but reliable as no variance is associated with it) & Good (often used along a simple threshold) & Poor (or rather constrained by how well outcomes follow a distribution) & Good (when using sampling over analytic forms) \\[0.25em]
Linear machine learning & Fair (regularization, especially, can yield parsimonious models and reduces the risk of overfitting) & Good (as coefficients directly quantify the effect size) & Poor (often not regarded or only interaction terms or predefined transformations such as logit) & Good (closed-form solution for ordinary least squares; optimization problems for regularization)  \\[0.25em]
Non-linear machine learning & Good (still regarded as the benchmark against which other models have to compete) & Poor (with the exception of certain approaches, \eg, random forests, that rank variable importance but still don't yield accountability of forecasts) & Good (can adapt well to subgroups, non-linear response curves and interactions) & Fair (often efficient estimations, but without uncertainty quantification) \\[0.25em]
Deep learning & Good (given sufficient training data) & Poor & Good (even when taking sequences as input) & Poor (challenging hyperparameter tuning) \\
\midrule
\textbf{Structured-effect neural network} & Good (theoretically identical to deep learning) & Good (full accountability of the structured effect) & Good (included but only confined to the variance that cannot be explained by the structured effect) & Fair (time-consuming sampling but less prone to unfavorable hyperparameters)\\
\bottomrule
\end{tabular}
}%
\caption{Stylized characteristics of different models in machine learning. Here we extend the categorization from \citet[p.\,351]{Hastie.2009} to include deep learning and our structured-effect neural network.}
\label{tbl:PredVsInterp}
\end{table} 

Sensors have always been an important part of predictive maintenance, as they allow to monitor and to adjust small changes so that small problems do not turn into big problems \citep[\eg][]{Rabatel.2011}. Many different sensors monitoring different measurements can be the key to better understanding processes and preventing early failures and consequent downtime. However, complex relationships between potentially large number of sensors and the effect on machinery demands for advanced, non-linear modeling of the remaining useful life. Thus, to fully exploit the information obtained from sensors, interpretability is of great use. Our structured-effect neural network bridges the gap between these key specifications.

\subsection{Limitations and potential for future research}

Recently, dropout as a Bayesian approximation has been proposed as a simple, yet efficient means to obtain uncertainty estimates for neural networks \citep{Gal.2016}. This approach leads to models with fewer parameters, which generally facilitates optimization. Further, computationally costs for optimization are lower, compared to the costs when utilizing variational Bayesian inference. However, Bayesian approximation via dropout does not provide uncertainty estimates for coefficients in our model. Additionally, Bayesian approximation via dropout comes at the cost of not being capable of including prior information about the coefficients into the model. As the latter is particularly important for predictive maintenance where expert knowledge is inevitable, we decided to utilize variational Bayesian inference.

\subsection{Concluding remarks}

Decision support as a field has developed a variety of approaches to improve the cost efficiency of maintenance, especially by predicting the remaining useful life of machinery and linking operational decision-making to it. Common approaches for predictive maintenance include statistical models based on probability density functions or machine learning, which further incorporates sensor data. While the former still serves as widespread common practice due to its reliability and interpretability, the latter has shown considerable improvements in prediction accuracy.

This research develops a new model that combines both advantages. Our suggested structured-effect neural network achieves accountability similar to simple distribution-based RUL models as its primary component, as well as a linear combination of sensor measurements. The remaining variance is then described by a recurrent neural network from the field of deep learning, which is known for its flexibility in adapting to non-linear relationships. For this purpose, all parameters are modeled as latent variables and we propose variational Bayesian inferences for their estimation in order to optimize the Kullback-Leibler divergence. Our findings reveal that our structured-effect neural network outperforms traditional machine learning models and still allows one to draw interpretable conclusions about the sources of the deterioration process. 

\appendix
\section{Derivation of ELBO for structured-effect neural network}
\label{sec:derivation}

Our suggested approach draws upon variational Bayesian methods and approximates the true posterior via a variational distribution $Q_\lambda(\theta) \approx P(\theta \,|\, X)$. Here $Q_\lambda(\theta)$ refers to a family of distributions that is indexed by $\lambda$ and, hence, our optimization problem translates into finding the optimal $\lambda^\ast$ along with the corresponding distribution $Q_{\lambda^\ast}$. The following theorems state the mathematical definition of $\lambda^\ast$ and introduce a tractable approximation.  

\begin{theorem}
	\label{thr:lambda_star}
	The optimal $\lambda^\ast$ is given by 
	\begin{equation}
	\lambda^\ast = \argmin_{\lambda} \; \expectation_{Q_\lambda} [ \log Q_\lambda (\theta) ] - \expectation_{Q_\lambda} [ \log P(X, \theta) ] + \log P(X) .
	\end{equation}
	
	\begin{proof}
		The fit between the variational distribution $Q_\lambda(\theta)$ and the posterior distribution $P(\theta \,|\, X)$ can be measured by the Kullback-Leibler divergence. Hence, we yield 
		\begin{equation}
		\lambda^\ast = \argmin_{\lambda} \; \mathit{KL}(Q_\lambda (\theta) \,||\, P(\theta \,|\, X)) .
		\end{equation}
		Inserting the definition of the Kullback-Leibler divergence results into
		\begin{align}
		\lambda^\ast &= \argmin_{\lambda}{\expectation_{Q_\lambda} \; [ \log Q_\lambda (\theta) ] - \expectation_{Q_\lambda} [ \log P(\theta \,|\, X) ]} \\
		&= \argmin_{\lambda}{\expectation_{Q_\lambda} \; [ \log Q_\lambda (\theta) ] - \expectation_{Q_\lambda} [ \log P(X, \theta) ] + \log P(X)} .
		\end{align}
		\hfill\qed
	\end{proof}
\end{theorem}
\vspace{-0.5cm}

Unfortunately, \Cref{thr:lambda_star} is intractable, as it depends on the marginal likelihood of the model, $\log P(X)$. Therefore, the following theorem derives an approximation for the marginal likelihood of the model. 

\begin{theorem}
	\label{thr:ELBO}
	The marginal likelihood of the model $\log P(X)$ can be approximated by the evidence lower bound, $ELBO(\lambda)$, \ie,
	\begin{equation}
	\log P(X) \geq \expectation_{Q_\lambda} [ \log P (X , \theta) ] - \expectation_{Q_\lambda} [ \log Q_\lambda(\theta) ] = \textit{ELBO}(\lambda).
	\end{equation}
	
	\begin{proof}
		Utilizing Jensen's inequality, it holds that
		\begin{align}
		\log P(X) &= \log \int \! P(X , \theta)\, \text{d}\theta
		= \log \int \! P(X , \theta) \frac{Q_\lambda (\theta)}{Q_\lambda (\theta)} \, \text{d}\theta 
		= \log \expectation_{Q_\lambda} \left[ \frac{P(X , \theta)}{Q_\lambda(\theta)} \right] \\
		&\geq \expectation_{Q_\lambda} \left[ \log \frac{P(X , \theta)}{Q_\lambda(\theta)} \right] 
		= \expectation_{Q_\lambda} [ \log P(X , \theta) ] - \expectation_{Q_\lambda} [ \log q(\theta) ]. 
		\end{align}
		\hfill\qed
	\end{proof}
\end{theorem}

\begin{theorem}
	The optimal $\lambda^\ast$ can be approximated by
	\begin{equation}
	\lambda^\ast = \argmax_{\lambda} \textit{ELBO}(\lambda).
	\end{equation}
	
	\begin{proof}
		From \Cref{thr:lambda_star} and \Cref{thr:ELBO}, it immediately follows that
		\begin{align}
		\lambda^\ast = \argmin_{\lambda} \; \log P(X) - \textit{ELBO}(\lambda).
		\end{align}
		As $\log P(X)$ is constant with respect to $\lambda$, the value $\lambda^\ast$ can be approximated by maximizing $\textit{ELBO}(\lambda)$.
		\hfill\qed
	\end{proof}
\end{theorem}
\vspace{-0.5cm}

In order to optimize $\textit{ELBO}(\lambda)$, we utilize gradient descent with the gradients defined by
\begin{align}
\label{equ:grad}
\nabla_\lambda \textit{ELBO}(\lambda) &= \nabla_\lambda \expectation_{Q_\lambda} [ \log P(X , \theta) ] - \expectation_{Q_\lambda} [ \log q(\theta) ] \\
&= \expectation_{Q_\lambda} [ \nabla_\lambda \log q(\theta) \left( \log P(X , \theta) - \log q(\theta) \right) ]. 
\end{align}
We further utilize Monte Carlo integration to obtain the estimates of the $\textit{ELBO}(\lambda)$ and the gradient.

\section{Two-stage estimation}
\label{sec:two_stage_optimization}

Analogous to our $\mathit{SENN}$, we chose the non-parametric component $\lambda$ to follow a log-normal prior. The underlying distribution parameters were modeled as normal prior distributions, \ie, $a \sim \mathcal{N}(a_{\text{empirical}},\,1)$ and $b \sim \mathcal{N}(b_{\text{empirical}},\,1)$. The linear component $\beta$ was modeled such that the coefficients stem from normal prior distributions,\ie, $\beta_i \sim \mathcal{N}(0,\,10)$. The recurrent component $\mathit{RNN}_\Theta$ was implemented as a long short-term memory network with two layers containing \num{100} and \num{50} neurons, respectively. Formally, the estimation is specified by as follows:
\begin{itemize}
\item Stage 1:  $\lambda^\ast, \beta^\ast = \argmax_{\lambda, \beta} \; P(\lambda, \beta \,|\, X)$, \\
\item Stage 2: $\mathit{RNN}_\Theta^\ast = \argmax_{\mathit{RNN}_\Theta} \; P(\mathit{RNN}_\Theta \,|\, X, \lambda^\ast, \beta^\ast)$.
\end{itemize}

\setlength{\bibsep}{0.15\baselineskip}
{\sloppy\small
\setlength{\bibsep}{3pt}
\bibliographystyle{model1-num-names}
\bibliography{literature}

\begin{thebibliography}{59}
\expandafter\ifx\csname natexlab\endcsname\relax\def\natexlab#1{#1}\fi
\providecommand{\url}[1]{\texttt{#1}}
\providecommand{\href}[2]{#2}
\providecommand{\path}[1]{#1}
\providecommand{\DOIprefix}{doi:}
\providecommand{\ArXivprefix}{arXiv:}
\providecommand{\URLprefix}{URL: }
\providecommand{\Pubmedprefix}{pmid:}
\providecommand{\doi}[1]{\href{http://dx.doi.org/#1}{\path{#1}}}
\providecommand{\Pubmed}[1]{\href{pmid:#1}{\path{#1}}}
\providecommand{\bibinfo}[2]{#2}
\ifx\xfnm\relax \def\xfnm[#1]{\unskip,\space#1}\fi
\bibitem[{Heidergott and Farenhorst-Yuan(2010)}]{Heidergott.2010}
\bibinfo{author}{B.~Heidergott}, \bibinfo{author}{T.~Farenhorst-Yuan},
\newblock \bibinfo{title}{Gradient estimation for multicomponent maintenance
  systems with age-replacement policy},
\newblock \bibinfo{journal}{Operations Research} \bibinfo{volume}{58}
  (\bibinfo{year}{2010}) \bibinfo{pages}{706--718}.
\bibitem[{Groenevelt et~al.(1992)Groenevelt, Pintelon, and
  Seidmann}]{Groenevelt.1992}
\bibinfo{author}{H.~Groenevelt}, \bibinfo{author}{L.~Pintelon},
  \bibinfo{author}{A.~Seidmann},
\newblock \bibinfo{title}{Production lot sizing with machine breakdowns},
\newblock \bibinfo{journal}{Management Science} \bibinfo{volume}{38}
  (\bibinfo{year}{1992}) \bibinfo{pages}{104--123}.
\bibitem[{Dogramaci and Fraiman(2004)}]{Dogramaci.2004}
\bibinfo{author}{A.~Dogramaci}, \bibinfo{author}{N.~M. Fraiman},
\newblock \bibinfo{title}{Replacement decisions with maintenance under
  uncertainty: An imbedded optimal control model},
\newblock \bibinfo{journal}{Operations Research} \bibinfo{volume}{52}
  (\bibinfo{year}{2004}) \bibinfo{pages}{785--794}.
\bibitem[{Lee and Rosenblatt(1987)}]{Lee.1987}
\bibinfo{author}{H.~L. Lee}, \bibinfo{author}{M.~J. Rosenblatt},
\newblock \bibinfo{title}{Simultaneous determination of production cycle and
  inspection schedules in a production system},
\newblock \bibinfo{journal}{Management Science} \bibinfo{volume}{33}
  (\bibinfo{year}{1987}) \bibinfo{pages}{1125--1136}.
\bibitem[{Jardine et~al.(2006)Jardine, Lin, and Banjevic}]{Jardine.2006}
\bibinfo{author}{A.~K. Jardine}, \bibinfo{author}{D.~Lin},
  \bibinfo{author}{D.~Banjevic},
\newblock \bibinfo{title}{A review on machinery diagnostics and prognostics
  implementing condition-based maintenance},
\newblock \bibinfo{journal}{Mechanical Systems and Signal Processing}
  \bibinfo{volume}{20} (\bibinfo{year}{2006}) \bibinfo{pages}{1483--1510}.
\bibitem[{Heng et~al.(2009)Heng, Zhang, Tan, and Mathew}]{Heng.2009}
\bibinfo{author}{A.~Heng}, \bibinfo{author}{S.~Zhang}, \bibinfo{author}{A.~C.
  Tan}, \bibinfo{author}{J.~Mathew},
\newblock \bibinfo{title}{Rotating machinery prognostics: State of the art,
  challenges and opportunities},
\newblock \bibinfo{journal}{Mechanical Systems and Signal Processing}
  \bibinfo{volume}{23} (\bibinfo{year}{2009}) \bibinfo{pages}{724--739}.
\bibitem[{Si et~al.(2011)Si, Wang, Hu, and Zhou}]{Si.2011}
\bibinfo{author}{X.-S. Si}, \bibinfo{author}{W.~Wang}, \bibinfo{author}{C.-H.
  Hu}, \bibinfo{author}{D.-H. Zhou},
\newblock \bibinfo{title}{Remaining useful life estimation: A review on the
  statistical data driven approaches},
\newblock \bibinfo{journal}{European Journal of Operational Research}
  \bibinfo{volume}{213} (\bibinfo{year}{2011}) \bibinfo{pages}{1--14}.
\bibitem[{Baptista et~al.(2018)Baptista, Sankararaman, de~Medeiros, Nascimento,
  Prendinger, and Henriques}]{Baptista.2018}
\bibinfo{author}{M.~Baptista}, \bibinfo{author}{S.~Sankararaman},
  \bibinfo{author}{I.~P. de~Medeiros}, \bibinfo{author}{C.~Nascimento},
  \bibinfo{author}{H.~Prendinger}, \bibinfo{author}{E.~M. Henriques},
\newblock \bibinfo{title}{Forecasting fault events for predictive maintenance
  using data-driven techniques and {ARMA} modeling},
\newblock \bibinfo{journal}{Computers {\&} Industrial Engineering}
  \bibinfo{volume}{115} (\bibinfo{year}{2018}) \bibinfo{pages}{41--53}.
\bibitem[{Seera et~al.(2014)Seera, Lim, Nahavandi, and Loo}]{Seera.2014}
\bibinfo{author}{M.~Seera}, \bibinfo{author}{C.~P. Lim},
  \bibinfo{author}{S.~Nahavandi}, \bibinfo{author}{C.~K. Loo},
\newblock \bibinfo{title}{Condition monitoring of induction motors: A review
  and an application of an ensemble of hybrid intelligent models},
\newblock \bibinfo{journal}{Expert Systems with Applications}
  \bibinfo{volume}{41} (\bibinfo{year}{2014}) \bibinfo{pages}{4891--4903}.
\bibitem[{Breiman(2001)}]{Breiman.2001}
\bibinfo{author}{L.~Breiman},
\newblock \bibinfo{title}{Statistical modeling: The two cultures},
\newblock \bibinfo{journal}{Statistical Science} \bibinfo{volume}{16}
  (\bibinfo{year}{2001}) \bibinfo{pages}{199--231}.
\bibitem[{Jang(2019)}]{Jang.2019}
\bibinfo{author}{H.~Jang},
\newblock \bibinfo{title}{A decision support framework for robust {R \& D}
  budget allocation using machine learning and optimization},
\newblock \bibinfo{journal}{Decision Support Systems} \bibinfo{volume}{121}
  (\bibinfo{year}{2019}) \bibinfo{pages}{1--12}.
\bibitem[{Subramania and Khare(2011)}]{Subramania.2011}
\bibinfo{author}{H.~S. Subramania}, \bibinfo{author}{V.~R. Khare},
\newblock \bibinfo{title}{Pattern classification driven enhancements for
  human-in-the-loop decision support systems},
\newblock \bibinfo{journal}{Decision Support Systems} \bibinfo{volume}{50}
  (\bibinfo{year}{2011}) \bibinfo{pages}{460--468}.
\bibitem[{Delen et~al.(2013)Delen, Zaim, Kuzey, and Zaim}]{Delen.2013}
\bibinfo{author}{D.~Delen}, \bibinfo{author}{H.~Zaim},
  \bibinfo{author}{C.~Kuzey}, \bibinfo{author}{S.~Zaim},
\newblock \bibinfo{title}{A comparative analysis of machine learning systems
  for measuring the impact of knowledge management practices},
\newblock \bibinfo{journal}{Decision Support Systems} \bibinfo{volume}{54}
  (\bibinfo{year}{2013}) \bibinfo{pages}{1150--1160}.
\bibitem[{Reifsnider and Case(2002)}]{Reifsnider.2002}
\bibinfo{author}{K.~L. Reifsnider}, \bibinfo{author}{S.~W. Case},
  \bibinfo{title}{Damage tolerance and durability of material systems},
  \bibinfo{publisher}{{Wiley Interscience}}, \bibinfo{address}{New York, NY},
  \bibinfo{year}{2002}.
\bibitem[{Liu et~al.(2006)Liu, Stratman, and Mahadevan}]{LIU.2006}
\bibinfo{author}{Y.~Liu}, \bibinfo{author}{B.~Stratman},
  \bibinfo{author}{S.~Mahadevan},
\newblock \bibinfo{title}{Fatigue crack initiation life prediction of railroad
  wheels},
\newblock \bibinfo{journal}{International Journal of Fatigue}
  \bibinfo{volume}{28} (\bibinfo{year}{2006}) \bibinfo{pages}{747--756}.
\bibitem[{Papakostas et~al.(2010)Papakostas, Papachatzakis, Xanthakis,
  Mourtzis, and Chryssolouris}]{Papakostas.2010}
\bibinfo{author}{N.~Papakostas}, \bibinfo{author}{P.~Papachatzakis},
  \bibinfo{author}{V.~Xanthakis}, \bibinfo{author}{D.~Mourtzis},
  \bibinfo{author}{G.~Chryssolouris},
\newblock \bibinfo{title}{An approach to operational aircraft maintenance
  planning},
\newblock \bibinfo{journal}{Decision Support Systems} \bibinfo{volume}{48}
  (\bibinfo{year}{2010}) \bibinfo{pages}{604--612}.
\bibitem[{Lipton(2018)}]{Lipton.2018}
\bibinfo{author}{Z.~C. Lipton},
\newblock \bibinfo{title}{The mythos of model interpretability},
\newblock \bibinfo{journal}{Communications of the ACM} \bibinfo{volume}{61}
  (\bibinfo{year}{2018}) \bibinfo{pages}{36--43}.
\bibitem[{Rudin(2018)}]{Rudin.2018}
\bibinfo{author}{C.~Rudin},
\newblock \bibinfo{title}{Please stop explaining black box models for high
  stakes decisions},
\newblock in: \bibinfo{booktitle}{Conference on Neural Information Processing
  Systems}, \bibinfo{year}{2018}.
\bibitem[{Lou et~al.(2013)Lou, Caruana, Gehrke, and Hooker}]{Lou.2013}
\bibinfo{author}{Y.~Lou}, \bibinfo{author}{R.~Caruana},
  \bibinfo{author}{J.~Gehrke}, \bibinfo{author}{G.~Hooker},
\newblock \bibinfo{title}{Accurate intelligible models with pairwise
  interactions},
\newblock in: \bibinfo{booktitle}{SIGKDD International Conference on Knowledge
  Discovery and Data Mining}, \bibinfo{year}{2013}, pp.
  \bibinfo{pages}{623--631}.
\bibitem[{Saxena and Goebel(2008)}]{Saxena.2008}
\bibinfo{author}{A.~Saxena}, \bibinfo{author}{K.~Goebel},
\newblock \bibinfo{title}{C-mapss data set},
\newblock \bibinfo{journal}{NASA Ames Prognostics Data Repository}
  (\bibinfo{year}{2008}).
\bibitem[{Butcher et~al.(2013)Butcher, Verstraeten, Schrauwen, Day, and
  Haycock}]{Butcher.2013}
\bibinfo{author}{J.~B. Butcher}, \bibinfo{author}{D.~Verstraeten},
  \bibinfo{author}{B.~Schrauwen}, \bibinfo{author}{C.~R. Day},
  \bibinfo{author}{P.~W. Haycock},
\newblock \bibinfo{title}{Reservoir computing and extreme learning machines for
  non-linear time-series data analysis},
\newblock \bibinfo{journal}{Neural networks} \bibinfo{volume}{38}
  (\bibinfo{year}{2013}) \bibinfo{pages}{76--89}.
\bibitem[{Dong et~al.(2017)Dong, Li, and Sun}]{Dong.2017}
\bibinfo{author}{D.~Dong}, \bibinfo{author}{X.-Y. Li}, \bibinfo{author}{F.-Q.
  Sun},
\newblock \bibinfo{title}{Life prediction of jet engines based on
  {LSTM}-recurrent neural networks},
\newblock in: \bibinfo{booktitle}{Prognostics and System Health Management
  Conference}, \bibinfo{publisher}{IEEE}, \bibinfo{year}{2017}.
\bibitem[{Liao and Kottig(2014)}]{Liao.2014}
\bibinfo{author}{L.~Liao}, \bibinfo{author}{F.~Kottig},
\newblock \bibinfo{title}{Review of hybrid prognostics approaches for remaining
  useful life prediction of engineered systems, and an application to battery
  life prediction},
\newblock \bibinfo{journal}{IEEE Transactions on Reliability}
  \bibinfo{volume}{63} (\bibinfo{year}{2014}) \bibinfo{pages}{191--207}.
\bibitem[{Navarro and Rychlik(2010)}]{Navarro.2010}
\bibinfo{author}{J.~Navarro}, \bibinfo{author}{T.~Rychlik},
\newblock \bibinfo{title}{Comparisons and bounds for expected lifetimes of
  reliability systems},
\newblock \bibinfo{journal}{European Journal of Operational Research}
  \bibinfo{volume}{207} (\bibinfo{year}{2010}) \bibinfo{pages}{309--317}.
\bibitem[{Dong and He(2007)}]{Dong.2007}
\bibinfo{author}{M.~Dong}, \bibinfo{author}{D.~He},
\newblock \bibinfo{title}{Hidden semi-{M}arkov model-based methodology for
  multi-sensor equipment health diagnosis and prognosis},
\newblock \bibinfo{journal}{European Journal of Operational Research}
  \bibinfo{volume}{178} (\bibinfo{year}{2007}) \bibinfo{pages}{858--878}.
\bibitem[{Kim and Makis(2013)}]{Kim.2013}
\bibinfo{author}{M.~J. Kim}, \bibinfo{author}{V.~Makis},
\newblock \bibinfo{title}{Joint optimization of sampling and control of
  partially observable failing systems},
\newblock \bibinfo{journal}{Operations Research} \bibinfo{volume}{61}
  (\bibinfo{year}{2013}) \bibinfo{pages}{777--790}.
\bibitem[{Plitsos et~al.(2017)Plitsos, Repoussis, Mourtos, and
  Tarantilis}]{Plitsos.2017}
\bibinfo{author}{S.~Plitsos}, \bibinfo{author}{P.~P. Repoussis},
  \bibinfo{author}{I.~Mourtos}, \bibinfo{author}{C.~D. Tarantilis},
\newblock \bibinfo{title}{Energy-aware decision support for production
  scheduling},
\newblock \bibinfo{journal}{Decision Support Systems} \bibinfo{volume}{93}
  (\bibinfo{year}{2017}) \bibinfo{pages}{88--97}.
\bibitem[{Ghosh et~al.(2007)Ghosh, Sharman, {Raghav Rao}, and
  Upadhyaya}]{Ghosh.2007}
\bibinfo{author}{D.~Ghosh}, \bibinfo{author}{R.~Sharman},
  \bibinfo{author}{H.~{Raghav Rao}}, \bibinfo{author}{S.~Upadhyaya},
\newblock \bibinfo{title}{Self-healing systems: survey and synthesis},
\newblock \bibinfo{journal}{Decision Support Systems} \bibinfo{volume}{42}
  (\bibinfo{year}{2007}) \bibinfo{pages}{2164--2185}.
\bibitem[{Mazhar et~al.(2007)Mazhar, Kara, and Kabernick}]{MAZHAR.2007}
\bibinfo{author}{M.~Mazhar}, \bibinfo{author}{S.~Kara},
  \bibinfo{author}{H.~Kabernick},
\newblock \bibinfo{title}{Remaining life estimation of used components in
  consumer products: Life cycle data analysis by {W}eibull and artificial
  neural networks},
\newblock \bibinfo{journal}{Journal of Operations Management}
  \bibinfo{volume}{25} (\bibinfo{year}{2007}) \bibinfo{pages}{1184--1193}.
\bibitem[{Cox(1972)}]{Cox.1972}
\bibinfo{author}{Cox},
\newblock \bibinfo{title}{Regression models and life-tables},
\newblock \bibinfo{journal}{Journal of the Royal Statistical Society}
  \bibinfo{volume}{34} (\bibinfo{year}{1972}) \bibinfo{pages}{187--220}.
\bibitem[{Wang et~al.(2013)Wang, Wang, Fang, and Chau}]{Wang.2013}
\bibinfo{author}{Y.~Wang}, \bibinfo{author}{S.~Wang},
  \bibinfo{author}{Y.~Fang}, \bibinfo{author}{P.~Y. Chau},
\newblock \bibinfo{title}{Store survival in online marketplace: An empirical
  investigation},
\newblock \bibinfo{journal}{Decision Support Systems} \bibinfo{volume}{56}
  (\bibinfo{year}{2013}) \bibinfo{pages}{482--493}.
\bibitem[{Zihajehzadeh and Park(2016)}]{Zihajehzadeh.2016}
\bibinfo{author}{S.~Zihajehzadeh}, \bibinfo{author}{E.~J. Park},
\newblock \bibinfo{title}{Regression model-based walking speed estimation using
  wrist-worn inertial sensor},
\newblock \bibinfo{journal}{PLOS ONE} \bibinfo{volume}{11}
  (\bibinfo{year}{2016}) \bibinfo{pages}{e0165211}.
\bibitem[{Riad et~al.(2010)Riad, Elminir, and Elattar}]{Riad.2010}
\bibinfo{author}{A.~Riad}, \bibinfo{author}{H.~Elminir},
  \bibinfo{author}{H.~Elattar},
\newblock \bibinfo{title}{Evaluation of neural networks in the subject of
  prognostics as compared to linear regression model},
\newblock \bibinfo{journal}{International Journal of Engineering {\&}
  Technology} \bibinfo{volume}{10} (\bibinfo{year}{2010})
  \bibinfo{pages}{52--58}.
\bibitem[{Mosallam et~al.(2013)Mosallam, Medjaher, and
  Zerhouni}]{Mosallam.2013}
\bibinfo{author}{A.~Mosallam}, \bibinfo{author}{K.~Medjaher},
  \bibinfo{author}{N.~Zerhouni},
\newblock \bibinfo{title}{Nonparametric time series modelling for industrial
  prognostics and health management},
\newblock \bibinfo{journal}{The International Journal of Advanced Manufacturing
  Technology} \bibinfo{volume}{69} (\bibinfo{year}{2013})
  \bibinfo{pages}{1685--1699}.
\bibitem[{Babu et~al.(2016)Babu, Zhao, and Li}]{SateeshBabu.2016}
\bibinfo{author}{G.~S. Babu}, \bibinfo{author}{P.~Zhao}, \bibinfo{author}{X.-L.
  Li},
\newblock \bibinfo{title}{Deep convolutional neural network based regression
  approach for estimation of remaining useful life},
\newblock in: \bibinfo{booktitle}{Database Systems for Advanced Applications},
  volume \bibinfo{volume}{9642}, \bibinfo{publisher}{Springer},
  \bibinfo{year}{2016}, pp. \bibinfo{pages}{214--228}.
\bibitem[{Wu et~al.(2018)Wu, Yuan, Dong, Lin, and Liu}]{Wu.2018}
\bibinfo{author}{Y.~Wu}, \bibinfo{author}{M.~Yuan}, \bibinfo{author}{S.~Dong},
  \bibinfo{author}{L.~Lin}, \bibinfo{author}{Y.~Liu},
\newblock \bibinfo{title}{Remaining useful life estimation of engineered
  systems using vanilla {LSTM} neural networks},
\newblock \bibinfo{journal}{Neurocomputing} \bibinfo{volume}{275}
  (\bibinfo{year}{2018}) \bibinfo{pages}{167--179}.
\bibitem[{Zheng et~al.(2017)Zheng, Ristovski, Farahat, and Gupta}]{Zheng.2017}
\bibinfo{author}{S.~Zheng}, \bibinfo{author}{K.~Ristovski},
  \bibinfo{author}{A.~Farahat}, \bibinfo{author}{C.~Gupta},
\newblock \bibinfo{title}{Long short-term memory network for remaining useful
  life estimation},
\newblock in: \bibinfo{booktitle}{IEEE International Conference on Prognostics
  and Health Management}, \bibinfo{publisher}{IEEE}, \bibinfo{year}{2017}, pp.
  \bibinfo{pages}{88--95}.
\bibitem[{Xiang et~al.(2018)Xiang, Musau, Wild, Lopez, Hamilton, Yang,
  Rosenfeld, and Johnson}]{Xiang.2018}
\bibinfo{author}{W.~Xiang}, \bibinfo{author}{P.~Musau}, \bibinfo{author}{A.~A.
  Wild}, \bibinfo{author}{D.~M. Lopez}, \bibinfo{author}{N.~Hamilton},
  \bibinfo{author}{X.~Yang}, \bibinfo{author}{J.~Rosenfeld},
  \bibinfo{author}{T.~T. Johnson},
\newblock \bibinfo{title}{Verification for machine learning, autonomy, and
  neural networks survey},
\newblock \bibinfo{journal}{arXiv preprint arXiv:1810.01989}
  (\bibinfo{year}{2018}).
\bibitem[{Athalye et~al.(2018)Athalye, Carlini, and Wagner}]{Athalye.2018}
\bibinfo{author}{A.~Athalye}, \bibinfo{author}{N.~Carlini},
  \bibinfo{author}{D.~Wagner},
\newblock \bibinfo{title}{Obfuscated gradients give a false sense of security:
  Circumventing defenses to adversarial examples},
\newblock in: \bibinfo{booktitle}{International Conference on Machine
  Learning}, \bibinfo{year}{2018}.
\bibitem[{Singh et~al.(2018)Singh, Gehr, Mirman, P{\"u}schel, and
  Vechev}]{Singh.2018}
\bibinfo{author}{G.~Singh}, \bibinfo{author}{T.~Gehr},
  \bibinfo{author}{M.~Mirman}, \bibinfo{author}{M.~P{\"u}schel},
  \bibinfo{author}{M.~Vechev},
\newblock \bibinfo{title}{Fast and effective robustness certification},
\newblock in: \bibinfo{booktitle}{Conference on Neural Information Processing
  Systems}, \bibinfo{year}{2018}, pp. \bibinfo{pages}{10802--10813}.
\bibitem[{Ribeiro et~al.(2016)Ribeiro, Singh, and Guestrin}]{Ribeiro.2016}
\bibinfo{author}{M.~T. Ribeiro}, \bibinfo{author}{S.~Singh},
  \bibinfo{author}{C.~Guestrin},
\newblock \bibinfo{title}{Why should i trust you? explaining the predictions of
  any classifier},
\newblock in: \bibinfo{booktitle}{SIGKDD International Conference on Knowledge
  Discovery and Data Mining}, \bibinfo{year}{2016}, pp.
  \bibinfo{pages}{1135--1144}.
\bibitem[{Lundberg and Lee(2017)}]{Lundberg.2017}
\bibinfo{author}{S.~M. Lundberg}, \bibinfo{author}{S.-I. Lee},
\newblock \bibinfo{title}{A unified approach to interpreting model
  predictions},
\newblock in: \bibinfo{booktitle}{Advances in Neural Information Processing
  Systems}, \bibinfo{year}{2017}, pp. \bibinfo{pages}{4765--4774}.
\bibitem[{{van der Maaten} and Hinton(2008)}]{vanderMaaten.2008}
\bibinfo{author}{L.~{van der Maaten}}, \bibinfo{author}{G.~Hinton},
\newblock \bibinfo{title}{Visualizing data using t-{SNE}},
\newblock \bibinfo{journal}{Journal of Machine Learning Research}
  \bibinfo{volume}{9} (\bibinfo{year}{2008}) \bibinfo{pages}{2579--2605}.
\bibitem[{Murdoch et~al.(2019)Murdoch, Singh, Kumbier, Abbasi-Asl, and
  Yu}]{Murdoch.2019}
\bibinfo{author}{W.~J. Murdoch}, \bibinfo{author}{C.~Singh},
  \bibinfo{author}{K.~Kumbier}, \bibinfo{author}{R.~Abbasi-Asl},
  \bibinfo{author}{B.~Yu},
\newblock \bibinfo{title}{Interpretable machine learning: Definitions, methods,
  and applications},
\newblock \bibinfo{journal}{arXiv preprint}  (\bibinfo{year}{2019}).
\bibitem[{Vlok et~al.(2002)Vlok, Coetzee, Banjevic, Jardine, and
  Makis}]{Vlok.2002}
\bibinfo{author}{P.~J. Vlok}, \bibinfo{author}{J.~L. Coetzee},
  \bibinfo{author}{D.~Banjevic}, \bibinfo{author}{A.~K.~S. Jardine},
  \bibinfo{author}{V.~Makis},
\newblock \bibinfo{title}{Optimal component replacement decisions using
  vibration monitoring and the proportional-hazards model},
\newblock \bibinfo{journal}{Journal of the Operational Research Society}
  \bibinfo{volume}{53} (\bibinfo{year}{2002}) \bibinfo{pages}{193--202}.
\bibitem[{Hastie et~al.(2009)Hastie, Tibshirani, and Friedman}]{Hastie.2009}
\bibinfo{author}{T.~Hastie}, \bibinfo{author}{R.~Tibshirani},
  \bibinfo{author}{J.~H. Friedman}, \bibinfo{title}{The Elements of Statistical
  Learning: Data Mining, Inference, and Prediction}, \bibinfo{edition}{2nd}
  ed., \bibinfo{publisher}{Springer}, \bibinfo{address}{New York, NY},
  \bibinfo{year}{2009}.
\bibitem[{Goodfellow et~al.(2017)Goodfellow, Bengio, and
  Courville}]{Goodfellow.2017}
\bibinfo{author}{I.~Goodfellow}, \bibinfo{author}{Y.~Bengio},
  \bibinfo{author}{A.~Courville}, \bibinfo{title}{Deep Learning},
  \bibinfo{publisher}{{MIT Press}}, \bibinfo{address}{Cambridge, MA},
  \bibinfo{year}{2017}.
\bibitem[{Hochreiter and Schmidhuber(1997)}]{Hochreiter.1997}
\bibinfo{author}{S.~Hochreiter}, \bibinfo{author}{J.~Schmidhuber},
\newblock \bibinfo{title}{Long short-term memory},
\newblock \bibinfo{journal}{Neural Computation} \bibinfo{volume}{9}
  (\bibinfo{year}{1997}) \bibinfo{pages}{1735--1780}.
\bibitem[{Fischer and Krauss(2018)}]{Fischer.2018}
\bibinfo{author}{T.~Fischer}, \bibinfo{author}{C.~Krauss},
\newblock \bibinfo{title}{Deep learning with long short-term memory networks
  for financial market predictions},
\newblock \bibinfo{journal}{European Journal of Operational Research}
  \bibinfo{volume}{270} (\bibinfo{year}{2018}) \bibinfo{pages}{654--669}.
\bibitem[{Kraus et~al.(2018)Kraus, Feuerriegel, and Oztekin}]{Kraus.2018}
\bibinfo{author}{M.~Kraus}, \bibinfo{author}{S.~Feuerriegel},
  \bibinfo{author}{A.~Oztekin},
\newblock \bibinfo{title}{Deep learning in business analytics and operations
  research: Models, applications and managerial implications},
\newblock \bibinfo{journal}{arXiv preprint}  (\bibinfo{year}{2018}).
\bibitem[{Srivastava and Lessmann(2018)}]{Srivastava.2018}
\bibinfo{author}{S.~Srivastava}, \bibinfo{author}{S.~Lessmann},
\newblock \bibinfo{title}{A comparative study of {LSTM} neural networks in
  forecasting day-ahead global horizontal irradiance with satellite data},
\newblock \bibinfo{journal}{Solar Energy} \bibinfo{volume}{162}
  (\bibinfo{year}{2018}) \bibinfo{pages}{232--247}.
\bibitem[{Tibshirani(1996)}]{Tibshirani.1996}
\bibinfo{author}{R.~Tibshirani},
\newblock \bibinfo{title}{Regression shrinkage and selection via the lasso},
\newblock \bibinfo{journal}{Journal of the Royal Statistical Society. Series B
  (Methodological)} \bibinfo{volume}{58} (\bibinfo{year}{1996})
  \bibinfo{pages}{267--288}.
\bibitem[{Ramasso(2014)}]{Ramasso.2014}
\bibinfo{author}{E.~Ramasso},
\newblock \bibinfo{title}{Investigating computational geometry for failure
  prognostics},
\newblock \bibinfo{journal}{International Journal of Prognostics and Health
  Management} \bibinfo{volume}{5} (\bibinfo{year}{2014})
  \bibinfo{pages}{1--18}.
\bibitem[{Bring(1994)}]{Bring.1994}
\bibinfo{author}{J.~Bring},
\newblock \bibinfo{title}{How to standardize regression coefficients},
\newblock \bibinfo{journal}{The American Statistician} \bibinfo{volume}{48}
  (\bibinfo{year}{1994}) \bibinfo{pages}{209--213}.
\bibitem[{Evermann et~al.(2017)Evermann, Rehse, and Fettke}]{Evermann.2017}
\bibinfo{author}{J.~Evermann}, \bibinfo{author}{J.-R. Rehse},
  \bibinfo{author}{P.~Fettke},
\newblock \bibinfo{title}{Predicting process behaviour using deep learning},
\newblock \bibinfo{journal}{Decision Support Systems} \bibinfo{volume}{100}
  (\bibinfo{year}{2017}) \bibinfo{pages}{129--140}.
\bibitem[{Kraus and Feuerriegel(2017)}]{Kraus.2017}
\bibinfo{author}{M.~Kraus}, \bibinfo{author}{S.~Feuerriegel},
\newblock \bibinfo{title}{Decision support from financial disclosures with deep
  neural networks and transfer learning},
\newblock \bibinfo{journal}{Decision Support Systems} \bibinfo{volume}{104}
  (\bibinfo{year}{2017}) \bibinfo{pages}{38--48}.
\bibitem[{Mahmoudi et~al.(2018)Mahmoudi, Docherty, and Moscato}]{Mahmoudi.2018}
\bibinfo{author}{N.~Mahmoudi}, \bibinfo{author}{P.~Docherty},
  \bibinfo{author}{P.~Moscato},
\newblock \bibinfo{title}{Deep neural networks understand investors better},
\newblock \bibinfo{journal}{Decision Support Systems} \bibinfo{volume}{112}
  (\bibinfo{year}{2018}) \bibinfo{pages}{23--34}.
\bibitem[{Rabatel et~al.(2011)Rabatel, Bringay, and Poncelet}]{Rabatel.2011}
\bibinfo{author}{J.~Rabatel}, \bibinfo{author}{S.~Bringay},
  \bibinfo{author}{P.~Poncelet},
\newblock \bibinfo{title}{Anomaly detection in monitoring sensor data for
  preventive maintenance},
\newblock \bibinfo{journal}{Expert Systems with Applications}
  \bibinfo{volume}{38} (\bibinfo{year}{2011}) \bibinfo{pages}{7003--7015}.
\bibitem[{Gal and Ghahramani(2016)}]{Gal.2016}
\bibinfo{author}{Y.~Gal}, \bibinfo{author}{Z.~Ghahramani},
\newblock \bibinfo{title}{Dropout as a bayesian approximation: Representing
  model uncertainty in deep learning},
\newblock in: \bibinfo{booktitle}{International Conference on International
  Conference on Machine Learning}, \bibinfo{year}{2016}, pp.
  \bibinfo{pages}{1050--1059}.

\end{thebibliography}
}

\end{document}